\documentclass[a4paper,num]{cas-dc}

\usepackage[numbers,sort&compress]{natbib}

\usepackage{enumitem} 

\let\printorcid\relax

\usepackage{tikz}
\newcommand{\circled}[1]{\tikz[baseline=(char.base)]{
    \node[shape=circle,draw,inner sep=0.5pt] (char) {\small #1};}}

\usepackage{algorithm}
\usepackage{algpseudocode}
\usepackage{amsmath}
\usepackage{amsfonts}

\usepackage{hyperref} 
\usepackage{cleveref} 

\crefname{equation}{Eq.}{Eqs.}  
\Crefname{equation}{Eq.}{Eqs.}  

\usepackage{enumitem}
\usepackage{subcaption}

\usepackage{tabularx} 
\usepackage{booktabs} 
\usepackage{siunitx}  
\usepackage{array}
\newcolumntype{Y}{>{\centering\arraybackslash\vcenter{\hbox\bgroup}}X<{\egroup}}

\def\tsc#1{\csdef{#1}{\textsc{\lowercase{#1}}\xspace}}
\tsc{WGM}
\tsc{QE}
\tsc{EP}
\tsc{PMS}
\tsc{BEC}
\tsc{DE}

\begin{document}
\let\WriteBookmarks\relax
\def\floatpagepagefraction{1}
\def\textpagefraction{.001}

\shorttitle{Expert Knowledge-driven Reinforcement Learning for Autonomous Racing via Trajectory Guidance and Dynamics Constraints}

\shortauthors{Bo Leng et~al.}

\title [mode = title]{Expert Knowledge-driven Reinforcement Learning for Autonomous Racing via Trajectory Guidance and Dynamics Constraints}                      



%




\author[1]{Bo Leng}[bioid = 1]
\credit{Conceptualization, Funding acquisition, Formal analysis, Project administration, Writing - review \& editing}

\author[1]{Weiqi Zhang}[bioid = 2]
\credit{Conceptualization, Data curation, Formal analysis, Investigation, Methodology, Software, Supervision, Validation, Visualization, Writing - original draft, Writing - review \& editing}

\author[1]{Zhuoren Li}[bioid = 3]
\cormark[1]
\credit{Conceptualization, Formal analysis, Methodology, Project administration, Supervision, Writing - original draft, Writing - review \& editing}

\author[1]{Lu Xiong}[bioid = 4]
\credit{Funding acquisition, Formal analysis, Resources, Supervision}

\author[1]{Guizhe Jin}[bioid = 5]
\credit{Methodology, Resources, Software, Supervision}

\author[1]{Ran Yu}[bioid = 6]
\credit{Methodology, Resources, Software, Visualization}

\author[2]{Chen Lv}[bioid = 7]
\credit{Conceptualization, Supervision, Writing - review \& editing}

\affiliation[1]{organization={College of Automotive and Energy Engineering, Tongji University},
    city={Shanghai},
    postcode={201804}, 
    country={China}}

\affiliation[2]{organization={School of Mechanical and Aerospace Engineering, Nanyang Technological University},
    city={Singapore},
    postcode={639798},
    country={Singapore}}


\cortext[cor1]{Corresponding author: Zhuoren Li (1911055@tongji.edu.cn)}



\begin{abstract}
Reinforcement learning has shown significant potential for autonomous racing, but it still faces challenges such as training instability, inefficient exploration, and unsafe action outputs in high-dynamic racing scenarios. This paper proposes a Trajectory guidance and Dynamics constraints Reinforcement Learning (TraD-RL) framework for autonomous racing. The proposed method incorporates expert prior knowledge into policy learning through Minimum Curvature Racing Line (MCRL) guidance, explicit vehicle dynamics constraints, and two-stage curriculum learning. MCRL provides global path and velocity references through observation augmentation and reward shaping, thereby improving exploration efficiency and racing performance. Yaw rate and sideslip angle constraints are introduced to characterize the vehicle dynamic safe operating envelope, and the corresponding stability costs are incorporated into policy optimization through Lagrangian relaxation. Moreover, the two-stage curriculum learning strategy enables a progressive transition from stable trajectory following to high-speed performance exploration. Experiments on two racetracks demonstrate that TraD-RL improves racing performance while maintaining a favorable balance between speed and dynamic stability. Further analyses of ablation, sensitivity, and robustness  validate the effectiveness and stability of the proposed framework.
\end{abstract}




\maketitle

\section{Introduction}

Autonomous racing has emerged as an important subfield of autonomous driving research. It features highly dynamic manoeuvres, strongly nonlinear vehicle behavior, and operating near the physical limits of handling. The primary objective is to minimize lap time while complying with race regulations and safety constraints. Autonomous race cars operate at high speeds and large accelerations, often with very little margin for error. This makes racing a demanding benchmark for decision-making and control methods that must remain reliable near the limits of feasibility. Moreover, it is also a quintessential safety-critical domain. Progress in this area drives decision-making and control methods that balance performance with safety and are suitable for verification and deployment. Recently, a growing body of work has examined planning, decision-making and control, and competitive interaction in autonomous racing \cite{muraleedharan2025randomized, kalaria2024alpha, langmann2025online}. International competitions such as the Indy Autonomous Challenge (IAC) \cite{betz2023tum, wischnewski2022indy} and the Abu Dhabi Autonomous Racing League (A2RL) \cite{a2rl2024, demeter2024lessons} have also attracted sustained participation from academia and industry, helping to drive rapid advances and keeping the field highly active.

Although traditional methods, such as Model Predictive Control (MPC) \cite{TIE2026}, have shown promising results in autonomous racing tasks, they still face limitations in terms of model accuracy, real-time performance and robustness \cite{bongard2025dynamic}. These methods typically divide the autonomous racing system into two modules: path planning and trajectory tracking. The planner generates an optimal trajectory based on vehicle and track information, while the controller follows that trajectory. However, in high-speed conditions, tires are often pushed to the friction limit, leading to highly nonlinear behavior \cite{brown2019coordinating}. Consequently, traditional optimization-based and rule-driven control strategies often struggle in these highly dynamic environments. Their rigid reliance on explicit mathematical formulations and predefined heuristics restricts the vehicle from fully exploiting its physical limits, leading to overly conservative driving behaviors and poor adaptability to unmodeled disturbances \cite{betz2022autonomous}.

As learning-based methods have demonstrated significant potential in high-dynamics control tasks \cite{song2021autonomous, kaufmann2023champion}, researchers have started to explore the use of learning-based decision-making and control in autonomous racing. This reduces reliance on precise modeling and enhances adaptability to strong nonlinearity and uncertainty. A representative approach is the incorporation of learning mechanisms into traditional MPC frameworks, leading to Learning-based MPC (LMPC). For example, \cite{kabzan2019learning} proposed a data-driven MPC controller based on online learning that uses Gaussian process regression to model residual uncertainties, allowing safer driving behaviors. \cite{xue2024learning} applied local linear, data-driven dynamic error learning to compensate for a nominal global nonlinear physical model, improving prediction accuracy and control performance under extreme operating conditions. Meanwhile, some studies have adopted imitation learning (IL) to learn racing strategies from expert or human demonstrations. For example, \cite{weaver2024betail} utilized human player demonstrations combined with adversarial imitation learning to improve the strategy's adaptability to new environments and distribution shifts. Despite the progress made by these learning-based methods in autonomous racing, LMPC still faces challenges related to model adaptation and parameter uncertainty, while imitation learning relies on high-quality expert demonstration datasets, which are difficult to obtain in practical applications.

Reinforcement Learning (RL) methods are more widely used in autonomous racing. By training agents through experience, the RL agent optimizes strategies based on reward feedback~\cite{zhu2026reinforcement}. RL does not require explicit model descriptions, making it particularly suited for tasks with high-dimensional state spaces and complex decision-making~\cite{TITS2026}. The core advantage of RL lies in maximizing long-term rewards, which aligns well with the goal of continuously optimizing lap times in racing tasks. RL has demonstrated performance that matches or exceeds human racers on various simulation platforms and even some physical systems. For instance, Fuch et al. \cite{fuchs2021super} utilized a course-progress proxy reward in Gran Turismo Sport, achieving lap times faster than both the in-game AI and over 50,000 human players. Wurman et al. \cite{wurman2022outracing} proposed a deep reinforcement learning approach that enabled an agent to outperform champion drivers in Gran Turismo Sport, showcasing RL’s potential in high-performance autonomous racing tasks. 

However, traditional RL methods typically require extensive interaction with the environment to optimize through trial and error, with low sample efficiency, high costs, and difficulty in iterative training in dynamic physical systems \cite{tang2025deep}. Additionally, in tasks with sparse rewards or narrow feasible regions, the lack of structured exploration guidance makes it challenging to obtain meaningful reward signals. This can result in slow convergence, unstable training, and the risk of ineffective exploration \cite{ladosz2022exploration}. More importantly, standard RL methods often neglect explicit modeling of safety constraints, failing to ensure their strict satisfaction. As a result, trial-and-error exploration during training may produce unsafe actions and violate constraints \cite{zhang2024safe}. Relying solely on safety rewards often fails to ensure strict constraint satisfaction, leading to unstable policy behavior and increased deployment risks. Recently, safe RL has been increasingly applied to autonomous driving policy learning to mitigate unsafe behaviors that may arise during the training and deployment of conventional reinforcement learning methods \cite{gu2024review}. However, existing safe RL studies in autonomous driving mainly focus on interaction safety between the ego vehicle and the external environment, such as collision avoidance, road-boundary constraints, and conflict risks with other traffic participants \cite{kalaria2023towards, NNLS2026}. To the best of our knowledge, no existing study has explicitly employed reinforcement learning to address the vehicle’s own dynamic stability safety in autonomous racing.

\begin{figure*}[!t]
	\centering
	\includegraphics[width=0.85\textwidth]{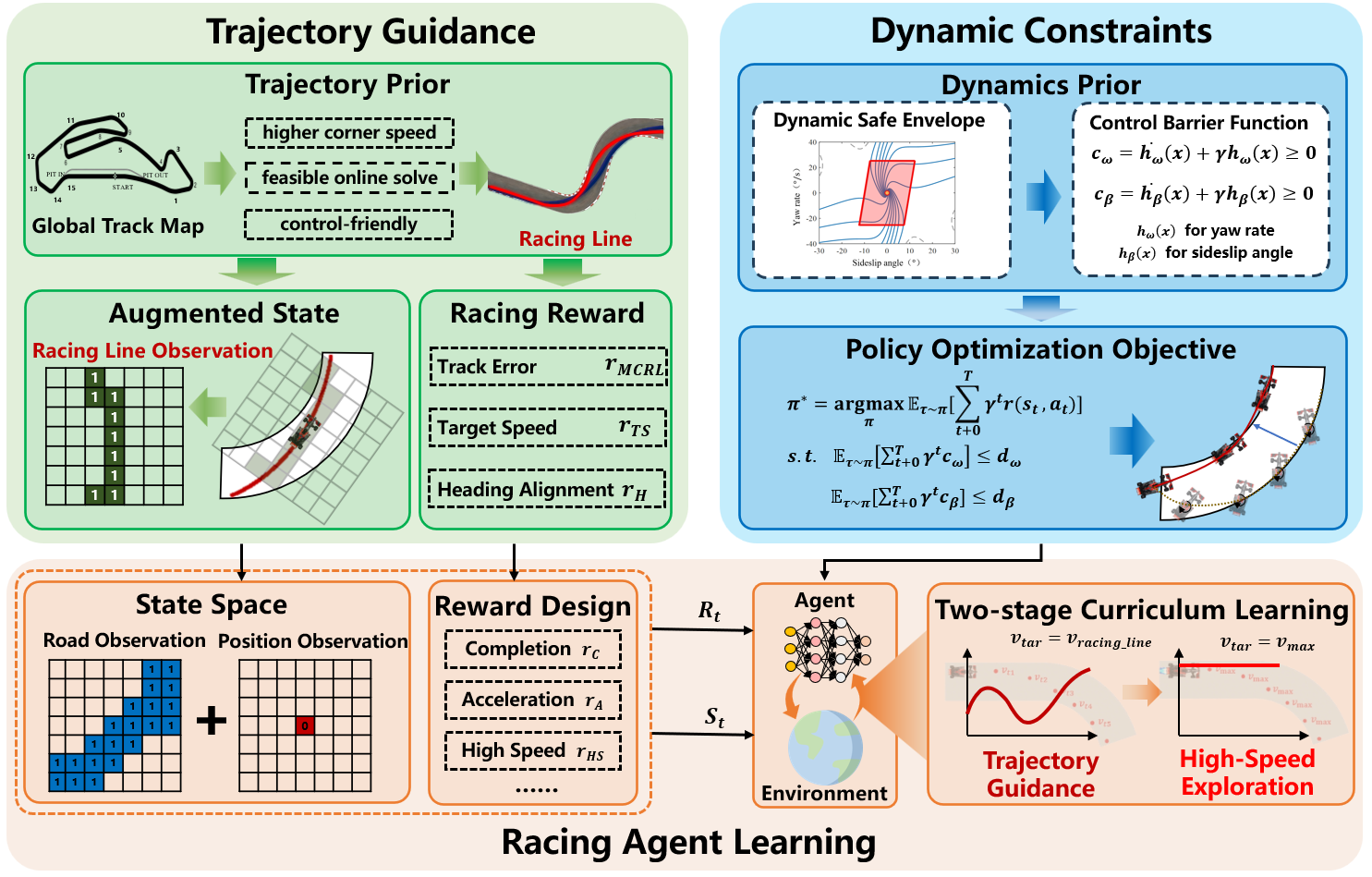} 
	\caption{The reinforcement learning decision-making and control framework driven by expert prior knowledge.}
	\label{fig:framework}
\end{figure*}

To address the aforementioned challenges, this paper proposes a \textbf{Tra}jectory guidance and \textbf{D}ynamics constraints \textbf{R}einforcement \textbf{L}earning (TraD-RL) decision-making and control framework for autonomous racing. Specifically, expert prior knowledge, encompassing trajectory guidance and dynamics constraints, is explicitly embedded within the training process. The overall framework is illustrated in Fig.~\ref{fig:framework}. First, to tackle the issues of low exploration efficiency and the difficulty of learning optimal trajectories in complex racing environments, we encode track geometric priors and racing intent into structured guidance information using the racing line. This approach enhances the learning efficiency and stability of the agent within high-dimensional spaces. Second, to reduce the risk of instability inherent in high-speed racing, we explicitly introduce vehicle dynamics stability constraints during policy learning. By comprehensively considering two key stability indicators, yaw rate and sideslip angle, this mechanism mitigates unsafe dynamic behaviors and improves deployment potential. Finally, to fully exploit the racing potential while ensuring training convergence, we develop a two-stage curriculum learning strategy based on dynamic task evolution. This facilitates a smooth transition of the model's capabilities toward limit-handling dynamic racing. The main contributions of this paper are as follows:
\begin{enumerate}[label=\arabic*), leftmargin=*, nosep]
    \item We propose an Minimum Curvature Racing Line (MCRL)-guided state representation and reward shaping mechanism for autonomous racing. The pre-computed MCRL is encoded into a reference-augmented observation space and used to construct dense rewards related to trajectory tracking, target-speed following, and heading alignment.  By constraining exploration around the reference priors, it alleviates the sparse-reward challenge in high-dimensional continuous action spaces and enables rapid convergence toward a high-performance policy that balances speed and stability;

    \item We introduce an explicit dynamics-constrained policy learning method based on vehicle dynamics priors. Yaw-rate and sideslip-angle limits are used to characterize a dynamics-based safe operating envelope, and their CBF-inspired safety margins are transformed into stability cost functions that incorporated into the policy optimization through Lagrangian relaxation. This mechanism regularizes unstable control behaviors during trial-and-error learning, thereby mitigating dynamic instability while maintaining sufficient exploration viability;
    
    \item We propose a progressive two-stage curriculum learning strategy. This strategy partitions the learning process into a trajectory guidance stage and a high-speed exploration stage. By implementing an 'easy-to-hard' training schedule through this progressive reward formulation, we significantly enhance both the learning efficiency and the racing performance of the reinforcement learning agent.
\end{enumerate}

The remainder of the paper is organized as follows: section \ref{sec:prelim} introduces preliminaries and defines the racing problem. The methodology is elaborated in Section \ref{sec:method}. In Section \ref{sec:results}, simulation testing and discussion of proposed method are presented. Section \ref{sec:conclusion} concludes this work.

\section{Preliminaries} \label{sec:prelim}

\subsection{Reinforcement Learning}

Reinforcement learning is typically mathematically formalized as a Markov Decision Process (MDP), defined by the tuple $\mathcal{M} = \langle \mathcal{O}, \mathcal{A}, \mathcal{P}, \mathcal{R}, \gamma \rangle$. Here, $\mathcal{O}$ denotes the state space, $\mathcal{A}$ represents the action space, $\mathcal{P} : \mathcal{O} \times \mathcal{A} \times \mathcal{O} \to [0,1]$ is the state transition probability function, $\mathcal{R} : \mathcal{O} \times \mathcal{A} \to \mathbb{R}$ is the reward function, and $\gamma \in [0,1)$ serves as the discount factor. At each discrete time step $t$, the agent observes the current state $o_t \in \mathcal{O}$ and selects an action $a_t \in \mathcal{A}$ according to a policy network $\pi_\theta(a_t | o_t)$, parameterized by $\theta$. The policy is typically constructed using neural networks (NNs) with the aim of fitting arbitrarily complex policy distribution functions. The primary objective of the agent is to determine an optimal policy $\pi^*_\theta$ that maximizes the expected cumulative discounted return $J(\pi_\theta)$:
\begin{equation}
    J(\pi_\theta) = \mathbb{E}_{\tau \sim \pi_\theta} \left[ \sum_{t=0}^{\infty} \gamma^t r(o_t, a_t) \right]
\end{equation}
where $\tau = (o_0, a_0, o_1, a_1, \dots)$ denotes the trajectory of state-action pairs.

To evaluate the performance of a policy, we define the state-value function $V^\pi(o)$ and the action-value function $Q^\pi(o, a)$. These functions satisfy the following Bellman expectation equations:

\begin{align}
    V^\pi(o_t) &= \mathbb{E}_{a_t \sim \pi, o_{t+1} \sim \mathcal{P}} \left[ r(o_t, a_t) + \gamma V^\pi(o_{t+1}) \right] \\
    Q^\pi(o_t, a_t) &= \mathbb{E}_{o_{t+1} \sim \mathcal{P}, a_{t+1} \sim \pi} \left[ r(o_t, a_t) + \gamma Q^\pi(o_{t+1}, a_{t+1}) \right]
\end{align}
where
$V^\pi(o_t)$ measures the long-term value of being in a specific state, while $Q^\pi(o_t, a_t)$ evaluates the value of taking a specific action from that state. Together, these equations serve as the theoretical foundation for value estimation and policy gradient computation.

In summary, the objective $J(\pi_\theta)$ provides the optimization goal for the agent, while the state-value and action-value functions serve as the essential metrics for evaluating the performance of the current policy $\pi_\theta$. The Bellman expectation equations establish the recursive consistency required to compute these value functions accurately. By solving these equations, the agent can effectively map its long-term objective into a series of local updates, enabling the improvement of the policy towards the global optimum~$\pi^*_\theta$.

\subsection{Vehicle Dynamics}

We employ a Dynamic Bicycle Model to capture lateral and yaw motion characteristics while maintaining computational efficiency. The vehicle is simplified as a single-mass rigid body on a two-dimensional plane. The state vector is defined as $X = [x, y, \varphi, u, v, \omega]^T$, where $(x, y)$ denotes the inertial position, $\varphi$ is the heading angle, $u$ and $v$ represent the longitudinal and lateral velocities in the body frame, and $\omega$ is the yaw rate.

Adopting the small angle approximation, the differential equations governing the vehicle dynamics are expressed as:
\begin{equation}
\begin{gathered}
    m(\dot{v} + u\omega) = F_{yf} \cos \delta + F_{yr} \\
    I_z \dot{\omega} = l_f F_{yf} \cos \delta - l_r F_{yr}
\end{gathered}
\end{equation}
where $m$ denotes the total vehicle mass, $I_z$ represents the yaw moment of inertia about the $z$-axis, and $l_f$ and $l_r$ are the distances from the center of mass to the front and rear axles, respectively.

From the differential equations, the system state-space representation, $\dot{X} = f(X, U)$, is derived as follows:

\begin{equation}
    \dot{X} = f(X, U) = \begin{bmatrix}
        u \cos \varphi - v \sin \varphi \\
        u \sin \varphi + v \cos \varphi \\
        \omega \\
        a + v \omega - \frac{1}{m} F_{yf} \sin \delta \\
        -u \omega + \frac{1}{m} (F_{yf} \cos \delta + F_{yr}) \\
        \frac{1}{I_z} (l_f F_{yf} \cos \delta - l_r F_{yr})
    \end{bmatrix}
\end{equation}
where $U = [a, \delta]^T$ is the control input, $a$ and $\delta$ represent longitudinal acceleration and the front-wheel steering angle, respectively.

To ensure both the efficiency and fidelity of the vehicle dynamic response, the interaction between the tires and the road is modeled within the linear elastic range. The lateral tire forces at the front and rear axles, $F_{yf}$ and $F_{yr}$, are proportional to their respective slip angles, $\alpha_f$ and $\alpha_r$, i.e., $F_{yf} = C_{\alpha f} \alpha_f$, $F_{yr} = C_{\alpha r} \alpha_r$,
where $C_{\alpha f}$ and $C_{\alpha r}$ denote the cornering stiffness of the front and rear tires, respectively. Based on the kinematic relationships, the slip angles are defined as:
\begin{equation}
\begin{aligned}
    \alpha_f &= \delta - \arctan \left( \frac{v + l_f \omega}{u} \right) \\
    \alpha_r &= - \arctan \left( \frac{v - l_r \omega}{u} \right)
\end{aligned}
\end{equation}

\section{Methodology} \label{sec:method}

The autonomous racing task is defined as a constrained time-optimal control problem and is formulated as MDP. The goal is to minimize lap time subject to strict vehicle dynamics and geometric track constraints. Within this formulation, a reinforcement learning agent learns to map local observations to continuous control inputs, specifically longitudinal acceleration and steering angle. This approach enables the derivation of a precise and safety-constrained policy through sequential interaction with the environment.

\subsection{RL Racing Framework}

\begin{figure}
	\centering
		\includegraphics[width=\columnwidth]{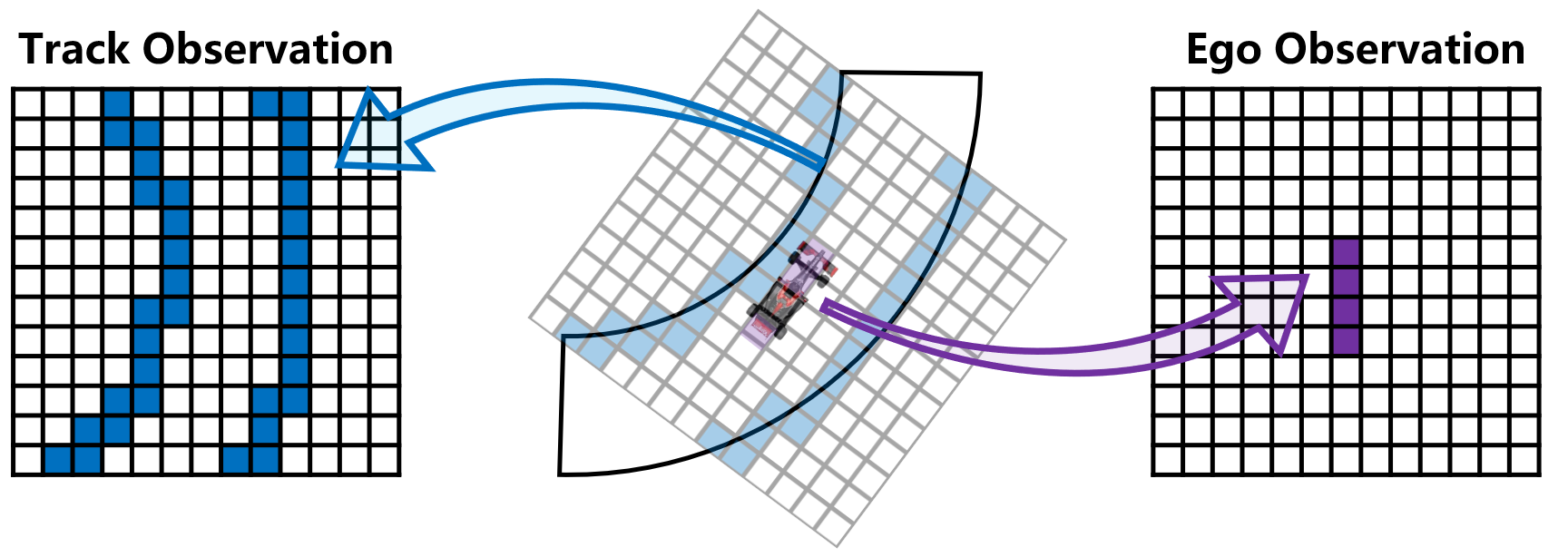}
	\caption{Schematic diagram of the ego-centric grid observation space construction.}
	\label{fig:obs_base}
\end{figure}

\textbf{Observation Space} This paper adopts an ego-centric occupancy grid map as the fundamental representation of the observation space. The observation vector consists of two parts: environmental geometric features and vehicle kinematic states, as illustrated in Fig.~\ref{fig:obs_base}. Specifically, the environmental features primarily consist of the track boundary observation $o_{\text{track}}$. The vehicle state observation includes the ego-vehicle's lateral velocity $o_{\text{ego\_v}}$, longitudinal velocity $o_{\text{ego\_u}}$, and heading angle $o_{\text{ego\_h}}$.

Regarding the feature encoding strategy, we employ a spatially aligned numerical mapping method: For track observation, binary encoding is used. Grid cells containing track boundary points take the value 1, while the others are 0. For ego-vehicle observation, numerical embedding encoding is used. The kinematic states of the ego-vehicle are inserted into the grid cells occupied by its geometric center. Furthermore, if expert prior knowledge is incorporated, the observation space will include the trajectory guidance information $o_{\text{MCRL}}$. The details are presented in \ref{obs_augment}. All observation components are represented as spatial feature maps with the same grid size and are concatenated along the channel dimension to form the final multi-channel observation tensor.

\textbf{Action Space} This paper constructs a continuous and decoupled two-dimensional action space. Specifically, at each time step, the reinforcement learning agent directly outputs the longitudinal acceleration command $a$ and the front-wheel steering angle command $\delta$. Furthermore, to enhance the convergence rate and numerical stability of the network training, all output actions are normalized to the interval $[-1, 1]$.

\textbf{Reward Function Design} To encourage the agent to complete the racing task as quickly as possible, this paper designs a multi-dimensional basic reward function. This function consists primarily of four components: the track constraint reward, the high-speed reward, the low-speed penalty, and the lap completion reward.

Track Constraint Reward: To ensure the race car operates within the drivable area, we define a track constraint reward: 
\begin{equation}
    r_{\text{track}} = 
    \begin{cases} 
        1 & \text{on track} \\
        -2 & \text{off track} 
    \end{cases}
\end{equation}

High-Speed Reward: When the speed falls within the high-speed range $[u_{h1}, u_{h2}]$, the reward increases linearly with speed. 
\begin{equation}
    r_{hs} = \frac{u - u_{h1}}{u_{h2} - u_{h1}}
\end{equation}

Low-Speed Penalty: When the speed falls within the low-speed range \([u_{l1}, u_{l2}]\), a low-speed penalty indicator is introduced to discourage the vehicle from remaining at excessively low speeds.
\begin{equation}
    r_{ls} = 1 - {(\frac{u}{u_{l2}})}^2
\end{equation}


Lap Completion Reward: To incentivize the agent to successfully complete a full lap, a discrete terminal reward $r_{\text{lap}}$ is granted whenever the vehicle crosses the finish line. 

\begin{equation}
    r_{\text{lap}} = 
    \begin{cases} 
      1, & \text{if } \text{a full lap is completed} \\
      0, & \text{otherwise}
    \end{cases}
\end{equation}

Finally, to eliminate magnitude discrepancies among different rewards and enhance numerical stability, all rewards designed in this paper are normalized to the interval $[-1,1]$.

\subsection{Trajectory Prior Guidance}

In the context of high-speed, highly dynamic autonomous racing, reinforcement learning often suffers from inefficient exploration due to the continuous action space and sparse environmental rewards. To address these challenges, this paper proposes a Trajectory Prior Guidance method, aimed at bridging the gap between random exploration and optimal control by incorporating expert knowledge. This expert trajectory is utilized not only to construct dense reward functions but also directly for observation space augmentation. This approach narrows the search range of the policy, thereby enhancing the training stability and convergence speed of the algorithm.

\subsubsection{Racing Line Generation}

The geometric features of the track can be characterized by a reference line. This line consists of a series of discrete global coordinates $[x_i, y_i]^T$, and includes the distance information to the left and right track boundaries for each point, denoted as $w_{left,i}$ and $w_{right,i}$, respectively. To generate the optimal racing line, we formulate the path planning task as a lateral offset optimization problem relative to the reference line, as illustrated in Fig.~\ref{fig:track_model}. Specifically, the $i$-th waypoint on the racing line $[x_{r,i}, y_{r,i}]^T$ is obtained by applying a lateral displacement along the normal vector $\vec{n}_i$ of the corresponding reference point:

\begin{figure}
	\centering
		\includegraphics[width=0.8\columnwidth]{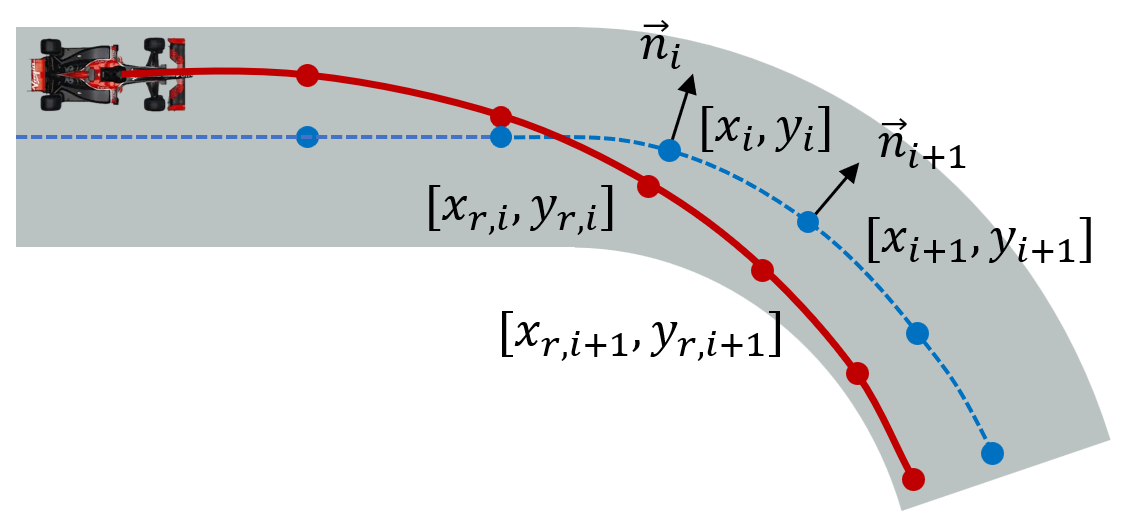}
	\caption{Parameterization of the optimal racing line relative to the track centerline.}
	\label{fig:track_model}
\end{figure}

\begin{equation}
    [x_{r,i}, y_{r,i}]^T = [x_i, y_i]^T + \sigma_i \vec{n}_i
\end{equation}
where $\sigma_i$ is the scalar parameter to be optimized, determining the lateral displacement distance of the racing line waypoint along the normal direction. To ensure the racing line always remains within the drivable area and avoids collisions, the value range of $\sigma_i$ is jointly determined by the track geometry and the vehicle width $w_{veh}$:
\begin{equation}
    \sigma_i \in \left[ -w_{left,i} + \frac{w_{veh}}{2}, w_{right,i} - \frac{w_{veh}}{2} \right]
\end{equation}

This paper defines the optimization objective as the minimization of the sum of squared curvatures along the discrete path:
\begin{equation} \label{equ:mcrl_slover}
\begin{aligned}
    \min_{\sigma_1, \dots, \sigma_N} \quad & \sum_{i=1}^{N} \kappa_i^2(t) \\
    \text{s.t.} \quad & \sigma_i \in \left[ -w_{left,i} + \frac{w_{veh}}{2}, w_{right,i} - \frac{w_{veh}}{2} \right]
\end{aligned}
\end{equation}

\begin{figure}
	\centering
		\includegraphics[width=1\columnwidth]{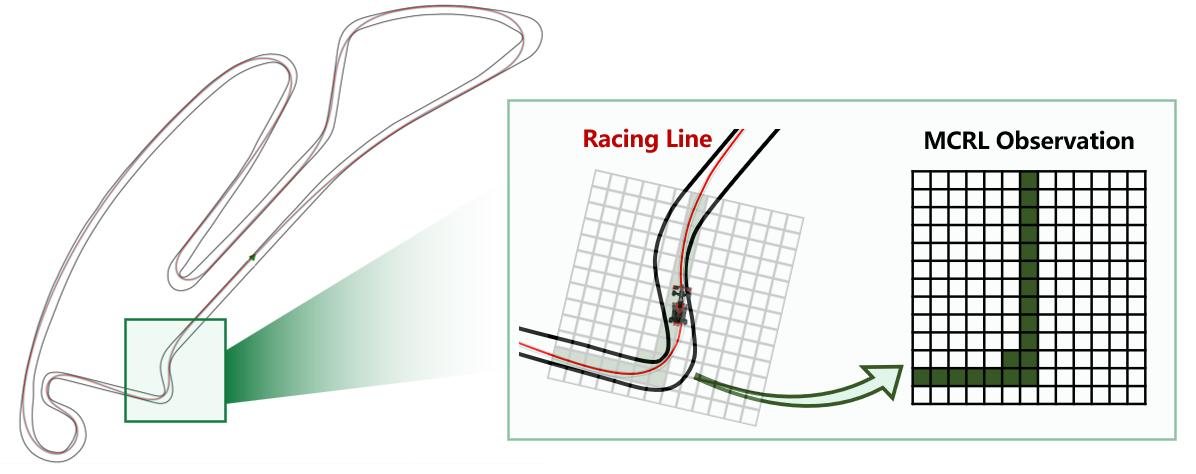}
	\caption{Illustration of the MCRL prior and its occupancy-grid encoding. The generated MCRL trajectory is first represented in the global racetrack map and then transformed into the vehicle-aligned local coordinate system, where it is encoded as the MCRL observation in the ego-centric occupancy grid.}
	\label{fig:racing_line_obs}
\end{figure}

An iterative optimization strategy is employed considering the non-linearity of curvature calculation and potential errors introduced by single-step linearization \cite{heilmeier2020minimum}. By iteratively refining the curvature until the difference between two adjacent iterations converges within a preset threshold $\varepsilon$ ($|\kappa_{n+1} - \kappa_n| \leq \varepsilon$), the racing line achieves higher geometric accuracy. Meanwhile, this process effectively eliminates local curvature peaks, yielding a trajectory that better aligns with the vehicle's kinematic characteristics. The Minimum Curvature Racing Line (MCRL) is illustrated in Fig.~\ref{fig:racing_line_obs}.

After obtaining the MCRL, the reference velocity profile is generated using a GGV-envelope-based forward--backward integration method \cite{kapania2016sequential}. Specifically, a speed-indexed GGV lookup table is constructed to provide the tire-level maximum longitudinal acceleration \(a_{x,\max}^{\mathrm{GGV}}\) and maximum lateral acceleration \(a_{y,\max}^{\mathrm{GGV}}\), which are both set to \(12~\mathrm{m/s^2}\) in this simulation setting. In addition, to account for the reduction in propulsion capability at high speeds, a speed-dependent machine-limited acceleration table \(a_{x,\max}^{\mathrm{mach}}(u)\) is introduced, which is only used to constrain the positive acceleration during the forward pass.

Given the curvature \(\kappa_i\) of the \(i\)-th waypoint on the MCRL, the curvature-limited longitudinal velocity is first calculated based on the lateral acceleration constraint:
\begin{equation}
u_i^{\mathrm{lat}}
=
\min
\left(
u_{\max},
\sqrt{
\frac{
a_{y,\max}^{\mathrm{GGV}}
}{
\max(|\kappa_i|,\varepsilon)
}
}
\right),
\end{equation}
where \(u_{\max}\) denotes the maximum longitudinal velocity, and \(\varepsilon\) is a small positive constant used to avoid division by zero.

A forward integration pass is then performed. For two adjacent waypoints \(i\) and \(i+1\), with the arc-length interval \(\Delta s_i\), the available positive longitudinal acceleration at waypoint \(i\) is defined as \(a_{x,i}^{+}=\min(a_{x,\max}^{\mathrm{GGV}},a_{x,\max}^{\mathrm{mach}}(u_i^{\mathrm{fwd}}))\). Based on this available acceleration, the forward velocity profile is updated as
\begin{equation}
u_{i+1}^{\mathrm{fwd}}
=
\min
\left(
u_{i+1}^{\mathrm{lat}},
\sqrt{
\left(u_i^{\mathrm{fwd}}\right)^2
+
2a_{x,i}^{+}\Delta s_i
}
\right).
\end{equation}

To ensure sufficient braking distance before entering high-curvature segments, a backward integration pass is further conducted from the end of the trajectory to the start. Under the simplified symmetric longitudinal acceleration envelope, the longitudinal acceleration limit in the GGV envelope is used to approximate the maximum braking deceleration magnitude, denoted as \(a_{x,i}^{-}=a_{x,\max}^{\mathrm{GGV}}\). The backward reachable velocity at waypoint \(i\) is calculated as
\begin{equation}
u_i^{\mathrm{bwd}}
=
\sqrt{
\left(u_{i+1}^{\mathrm{ref}}\right)^2
+
2a_{x,i}^{-}\Delta s_i
}.
\end{equation}

The final velocity profile is obtained by intersecting the forward and backward profiles: $u_i^{\mathrm{ref}}=\min\left(u_i^{\mathrm{fwd}}, 
u_i^{\mathrm{bwd}}\right)$. Finally, the expert prior trajectory is encapsulated into a six-dimensional state vector \([s,x,y,\varphi,\kappa,u]\), comprising the curvilinear distance \(s\), inertial coordinates \((x,y)\), heading angle \(\varphi\), path curvature \(\kappa\), and longitudinal velocity \(u\).

\subsubsection{Observation Augmentation} \label{obs_augment}

Based on the racing line data generated above, we augment the primary observation space by constructing the racing line observation feature $o_{\text{MCRL}}$. As illustrated in Fig.~\ref{fig:racing_line_obs}, this feature is constructed within the ego-centric local coordinate system and encoded using a binary occupancy grid. Specifically, grid cells covering the racing line waypoints take the value 1, while the remaining areas are 0. The incorporation of this expert prior information significantly enhances training efficiency. It not only assists the agent in rapidly comprehending the trajectory guidance mechanism but also effectively compresses the policy exploration space, enabling precise acquisition of the driving policy.

\subsubsection{Reward Shaping}

In order to utilize expert prior knowledge to guide the agent's policy optimization process, the reward function is further refined through reward shaping. Specifically, three core reward components are constructed based on trajectory tracking error, speed tracking error, and heading alignment relative to the MCRL.

\begin{itemize}[leftmargin=*, nosep]
    \item Trajectory Tracking Reward: This reward aims to guide the race car to strictly adhere to the optimal trajectory coordinates. Let $l$ denote the Euclidean distance between the vehicle's current position $[x, y]^T$ and the nearest reference point $[x_r, y_r]^T$ on the racing line.
    
    \begin{equation} \label{equ:r_mcrl}
        r_{\text{MCRL}} = \frac{1}{1 + l^2}
    \end{equation}

    \item Target Speed Tracking Reward: We set the target speed $u_{tar}$ to the reference speed at the look-ahead point using a preview mechanism. This reward guides the agent to acquire the optimal longitudinal speed control strategy adapted to different track segments.

    \begin{equation} \label{equ:r_ts}
        r_{TS}
        =
        \frac{
        u_{\mathrm{tar}}
        -
        \min \left(
        |u-u_{\mathrm{tar}}|,
        u_{\mathrm{tar}}
        \right)
        }{
        \max(u_{\mathrm{tar}},\varepsilon)
        }.
    \end{equation}

    \item Heading Alignment Reward: Similarly leveraging the preview mechanism, we designate the heading at the look-ahead point on the racing line as the target heading $\varphi_{tar}$. This reward incentivizes precise vehicle pose control. 

    \begin{equation} \label{equ:r_h}
        r_{H} = \frac{1}{1 + (\varphi - \varphi_{tar})^2}
    \end{equation}
\end{itemize}

\subsection{Dynamics Constraints}

\begin{figure}
	\centering
		\includegraphics[width=0.55\columnwidth]{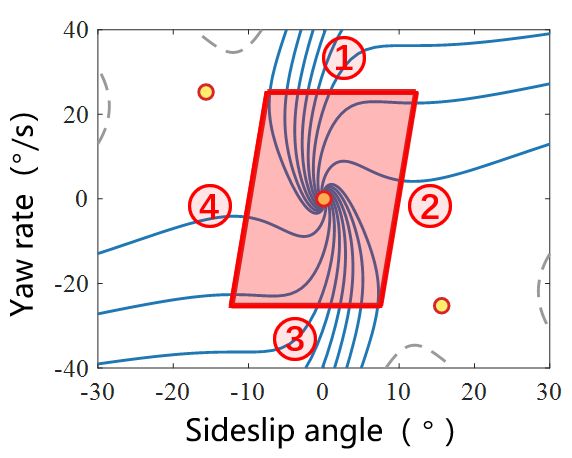}
	\caption{Illustration of the vehicle stability envelope based on the $\beta-\omega$ phase plane.}
	\label{fig:safe_envelope}
\end{figure}

The vehicle's stability region within the sideslip angle-yaw rate $\beta-\omega$ phase plane is characterized by incorporating explicit priors on vehicle dynamics control, thereby constructing an explicit safe operating envelope as illustrated in Fig.~\ref{fig:safe_envelope}. This approach confines the agent's policy updates within the dynamically feasible region, improving handling stability while the vehicle pursues limit performance.

As derived in \cite{zhu2023survey}, boundaries \circled{1} and \circled{3} in Fig.~\ref{fig:safe_envelope} define the yaw rate limits with no longitudinal sliding:
\begin{equation}
    |\omega_{ss,\lim}| = \frac{\mu mg}{mu} = \frac{\mu g}{u}
\end{equation}

Furthermore, the sideslip angle limits, boundaries \circled{2} and \circled{4}, are defined as:
\begin{equation}
    \beta_{\lim} = |\alpha_{r,\text{peak}}| + \frac{l_r \omega}{u}
\end{equation}
where $\alpha_{r,\text{peak}}$ is derived based on the Pacejka tire model \cite{ryu2013development}. By introducing the approximate limit slip angle as the saturation condition for rear tire force, the rear slip angle corresponding to the maximum lateral force is obtained as:

\begin{equation}
    \alpha_{r,\text{peak}} = \arctan \left( \frac{3 \mu m g l_f}{C_{\alpha r} (l_f + l_r)} \right)
\end{equation}

Based on the aforementioned safe operating envelope for yaw rate and sideslip angle, we adopt a CBF-inspired \cite{ames2019control} inequality form to represent the safety margins of the vehicle dynamics states. Specifically, a linear function is used to construct the following differentiable safety margin functions and the corresponding soft dynamics constraints:

\begin{itemize}[leftmargin=*, nosep]
    \item Yaw Rate Constraint:
    \begin{equation} \label{equ:yaw rate constraint}
        \dot{h}_{\omega} \geq -\alpha h_{\omega}, \quad h_{\omega} = \frac{\mu g}{u} - |\omega|
    \end{equation}

    \item Sideslip Angle Constraint:
    \begin{equation}
        \begin{aligned}
            \dot{h}_{\beta 1} \geq -\alpha h_{\beta 1}, \quad h_{\beta 1} = \beta_{\max} - \beta \\
            \dot{h}_{\beta 2} \geq -\alpha h_{\beta 2}, \quad h_{\beta 2} = \beta - \beta_{\min}
        \end{aligned}
    \end{equation}
\end{itemize}
where $\beta_{\min}$ and $\beta_{\max}$ correspond to the lower and upper bounds of the sideslip angle determined by boundaries \circled{2} and \circled{4} in Fig.~\ref{fig:safe_envelope}, respectively.

Accordingly, we define the instantaneous calculation formulas for the yaw rate cost $c_{\omega}$ and the sideslip angle cost $c_{\beta}$ as follows:

\begin{equation} \label{equ:c_omega}
    c_{\omega} = \max(-(\dot{h}_{\omega} + \alpha h_{\omega}), 0)
\end{equation}
\begin{equation} \label{equ:c_beta}
    c_{\beta} = \max(-\min(\dot{h}_{\beta 1} + \alpha h_{\beta 1}, \dot{h}_{\beta 2} + \alpha h_{\beta 2}), 0)
\end{equation}

Notably, this paper introduces a sliding-window mechanism in the actual policy iteration process. The window length remains fixed throughout training, and the sliding window is updated at each step. This mechanism approximates the degree of constraint violation by calculating the average cost within the most recent time window, which is then used for subsequent parameter updates. Compared to directly using instantaneous values, this strategy effectively mitigates the impact of data noise and single-step fluctuations on training stability.

To maximize task rewards while considering the aforementioned safety constraints, the RL problem is formulated as the following constrained optimization problem:
\begin{equation}
    \begin{aligned}
        \max_{\theta} \quad & J_{R}(\pi_{\theta}) = \mathbb{E}_{\tau \sim \pi_{\theta}} \left[ \sum_{t=0}^{T} \gamma^{t} r_{t} + \upsilon \mathcal{H}(\pi_{\theta}(\cdot \mid s_{t})) \right] \\
        \text{s.t.} \quad & \mathbb{E}_{\tau \sim \pi_{\theta}} [c_{\omega}(s, a)] \leq d_{\omega} \\
        & \mathbb{E}_{\tau \sim \pi_{\theta}} [c_{\beta}(s, a)] \leq d_{\beta}
    \end{aligned}
\end{equation}
where $\pi_{\theta}$ denotes the parameterized policy network, and $J_{R}(\pi_{\theta})$ represents the soft cumulative reward objective incorporating the entropy regularization term $\mathcal{H}$. $d_{\omega}$ and $d_{\beta}$ are the allowable cost thresholds.

Adaptive Lagrangian multipliers $\lambda_{\omega} \geq 0$ and $\lambda_{\beta} \geq 0$ are introduced to solve this problem, thereby transforming the constrained formulation into the following unconstrained optimization problem:

\begin{equation}
\begin{aligned}
    &\min_{\lambda_{\omega}, \lambda_{\beta} \geq 0} \max_{\theta}  L(\theta, \lambda_{\omega}, \lambda_{\beta}) \\
    &= J_{R}(\pi_{\theta}) - \sum_{k \in \{\omega, \beta\}} \lambda_{k} \left( \mathbb{E}_{(s, a) \sim \rho_{\pi_{\theta}}} [c_{k}(s, a)] - d_{k} \right)
\end{aligned}
\end{equation}

We adopt a primal-dual optimization method, performing alternating updates on the policy parameters $\theta$ and the Lagrangian multipliers $\lambda$: With the policy parameters $\theta$ fixed, the Lagrangian multipliers $\lambda$ are updated via gradient ascent to increase the penalty weights for constraint-violating behaviors; With the Lagrangian multipliers $\lambda$ fixed, the policy parameters $\theta$ are updated to maximize the aforementioned objective. In this process, the maximization of task rewards is subject to the regularization constraints imposed by the weighted constraint costs.

\subsection{Two-Stage Curriculum Design and RL Agent Training}

To balance convergence efficiency in the early stages of training with the maximization of performance potential in the later stages, this paper proposes a two-stage training strategy. This strategy partitions the entire training process into two distinct stages: "Trajectory Guidance" and "High-Speed Exploration," utilizing a preset training step threshold, $T_{switch}$, as the criterion for stage transition. This design enables the agent to graduate from trajectory imitation to the autonomous exploration of the vehicle's dynamic boundaries.

\begin{itemize}[leftmargin=*]
    \item Expert Trajectory Guidance Stage: The target speed within the target speed tracking reward function, $r_{TS}$, is designated as the reference speed at the corresponding point on the MCRL. This leverages the prior knowledge embedded in the MCRL to facilitate the agent's rapid mastery of fundamental racing maneuvers, such as heavy braking before corner entry and path maintenance at the apex. 

    \item High-Speed Exploration Stage: The target speed is reset to the maximum physically allowable speed of the racing vehicle. The removal of the MCRL speed constraints aims to overcome the conservatism inherent in the prior trajectory. It incentivizes the agent to explore within a broader state space, thereby uncovering faster lap times that surpass the MCRL baseline.
\end{itemize}

We employ Cost Critic networks, $Q_{C}$, for the accurate evaluation of potential constraint violation risks by estimating the expected cumulative constraint costs. For the $k$-th constraint, where $k \in \{\omega, \beta\}$, the parameters $\xi_{k}$ of the Cost Critic are updated by minimizing the following Mean Squared Error (MSE) loss function, $\mathcal{L}_{Cost}$:

\begin{equation} \label{equ:cost_c_loss}
    \mathcal{L}_{Cost}(\xi_{k}) = \mathbb{E}_{(s, a, c_{k}, s') \sim \mathcal{D}} \left[ \left( Q_{C_{k}}^{\xi} (s, a) - y_{c_{k}} \right)^{2} \right]
\end{equation}
where the target value $y_{c_{k}}$ is defined as:

\begin{equation}
    y_{c_{k}} = c_{k} + \gamma \max_{j=1,2} Q_{C_{k}, target}^{\xi_{j}} (s', a')
\end{equation}

The optimization objective of the Actor network is to maximize the reward while satisfying safety constraints. We integrate the constraint terms into the objective function and introduce a ReLU-based truncation mechanism, applying penalties only when the predicted cost exceeds the preset threshold $d_{k}$. This design creates a "safety dead-zone," allowing the agent to focus on task performance within the safe region and avoiding unnecessary conservative behavior. The loss function $\mathcal{L}_{Actor}$ for the actor is defined as:

\begin{equation} \label{equ:actor_loss}
\begin{aligned}
    \mathcal{L}_{Actor}(\theta) &= \mathbb{E}_{s \sim \mathcal{D}, a \sim \pi_{\theta}} \Biggl[ \underbrace{\alpha \log \pi_{\theta}(a \mid s) - \min_{i=1,2} Q_{\psi_{i}}(s, a)}_{\text{Reward Objective}} \\
    & + \sum_{k \in \{\omega, \beta\}} \lambda_{k} \cdot \underbrace{\text{ReLU} \left( \max_{j=1,2} Q_{C_{k}}^{\xi_{j}}(s, a) - d_{k} \right)}_{\text{Violation Penalty}} \Biggr]
\end{aligned}
\end{equation}
where $\psi$ denotes the parameters of the critic network.

The Lagrangian multipliers $\lambda_{k}$ are updated via Dual Ascent. The corresponding loss function $\mathcal{L}_{Lag}$, which aims to maximize the dual objective, is given by:

\begin{equation} \label{equ:lambda_loss}
    \mathcal{L}_{Lag}(\lambda_{k}) = -\lambda_{k} (\bar{C}_{k} - d_{k})
\end{equation}
where $\bar{C}_{k}$ is the average predicted cost of the current batch.

\begin{algorithm}[t]
\caption{TraD-RL Algorithm}
\label{alg:exprl}
\begin{algorithmic}[1]
\item[] \hspace*{-\algorithmicindent} \parbox[t]{\linewidth}{\textbf{Input:} Max training steps $T_{max}$, start training threshold $T_{start}$, stage switch threshold $T_{switch}$, safety cost thresholds $d_{\omega}, d_{\beta}$, batch size $N$.}
\item[] \hspace*{-\algorithmicindent} \parbox[t]{\linewidth}{\textbf{Initialize:} Randomly initialize Actor network $\pi_{\theta}$, Critic networks $Q_{\psi_{1,2}}$, safety Cost-Critic networks $Q_{\xi_{1,2}}^{\omega}, Q_{\xi_{1,2}}^{\beta}$; initialize Lagrangian multipliers $\lambda_{\omega}, \lambda_{\beta}$; initialize target network parameters $\psi', \xi'$; initialize experience replay buffer $\mathcal{D}$ and cost replay buffer $\mathcal{D}_{cost}$.}

\State Solve for MCRL according to \Cref{equ:mcrl_slover} based on the track map.
\For{current step $t = 1 \to T_{max}$}
    \State Obtain environmental observation $o_t$ including augmented observation $o_{MCRL}$.
    \State Select action $a_t \sim \pi_{\theta}(o_t)$ according to the current policy.
    \State Execute current action $a_t$, obtain basic rewards , obtain racing line related rewards according to \Cref{equ:r_mcrl,equ:r_ts,equ:r_h}, and calculate costs according to \Cref{equ:c_omega,equ:c_beta}.
    \State Store tuple $\langle o_t, a_t, r_t, o_{t+1} \rangle$ in $\mathcal{D}$; store tuple $\langle c_{\omega,t}, c_{\beta,t} \rangle$ in $\mathcal{D}_{cost}$.
    
    \If{$t > T_{start}$}
        \State Randomly sample $N$ tuples from $\mathcal{D}$ and $\mathcal{D}_{cost}$: 
        \Statex \hskip\algorithmicindent $\{ \langle o_i, a_i, r_i, o_{i+1} \rangle \}_{i=1 \dots N}$ and $\{ \langle c_{\omega,i}, c_{\beta,i} \rangle \}_{i=1 \dots N}$.
        \State Update Cost-Critic and Value networks using \Cref{equ:cost_c_loss}.
        \State Update Policy network using \Cref{equ:actor_loss}.
        \State Update Lagrangian multipliers using \Cref{equ:lambda_loss}.
        \State Update entropy regularization parameter $\upsilon$.
        \State Soft update target networks and target cost networks:
        \Statex \hskip\algorithmicindent $\psi'_1 \leftarrow \epsilon \psi_1 + (1-\epsilon)\psi'_1, \quad \psi'_2 \leftarrow \epsilon \psi_2 + (1-\epsilon)\psi'_2$
        \Statex \hskip\algorithmicindent $\xi'_1 \leftarrow \epsilon \xi_1 + (1-\epsilon)\xi'_1, \quad \xi'_2 \leftarrow \epsilon \xi_2 + (1-\epsilon)\xi'_2$
    \EndIf
\EndFor
\end{algorithmic}
\end{algorithm}

Specifically, the pseudocode for TraD-RL is presented in Algorithm \ref{alg:exprl}.

\section{Experiment Studies} \label{sec:results}

\subsection{Experiment Setup}

\begin{figure}
    \centering
    \begin{subfigure}[t]{0.48\columnwidth}
        \centering
        \includegraphics[width=\linewidth]{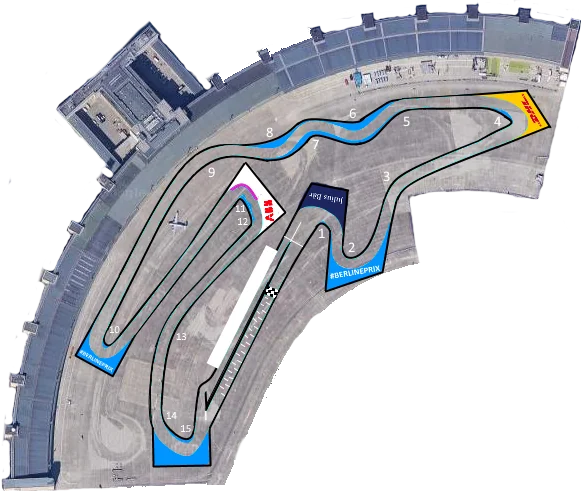}
        \caption{Berlin Racetrack}
        \label{fig:berlin_env}
    \end{subfigure}
    \hfill
    \begin{subfigure}[t]{0.48\columnwidth}
        \centering
        \includegraphics[width=\linewidth]{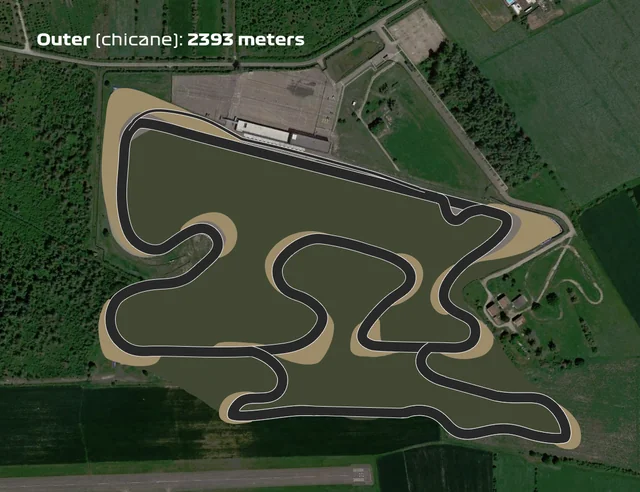}
        \caption{Modena Racetrack}
        \label{fig:modena_env}
    \end{subfigure}
    \caption{Aerial views of the two racetracks used in the simulation environments.}
    \label{fig:real_env}
\end{figure}

\textbf{Environment Setup} The simulation environments are constructed based on two racetracks: the Berlin Tempelhof Airport Street Circuit and the Modena Racetrack, as illustrated in Fig.~\ref{fig:real_env}. The Berlin track is a 2.469 km Formula E circuit with 17 turns~\cite{fia_berlin_circuit}, while the Modena track adopts the Modena 2019 configuration of the Autodromo di Modena in Italy~\cite{autodromo_modena_circuit}. In this work, the agent is trained on the Berlin Racetrack and evaluated on both the Berlin and Modena Racetracks.

\textbf{Algorithm Setup} The detailed experimental parameters are as follows. All training results are reported as the average over five random seeds, while all testing experiments are conducted under the same evaluation seed to ensure consistency.

\begin{figure}
    \centering
    \begin{subfigure}{\columnwidth} 
        \centering
        \includegraphics[width=0.9\linewidth]{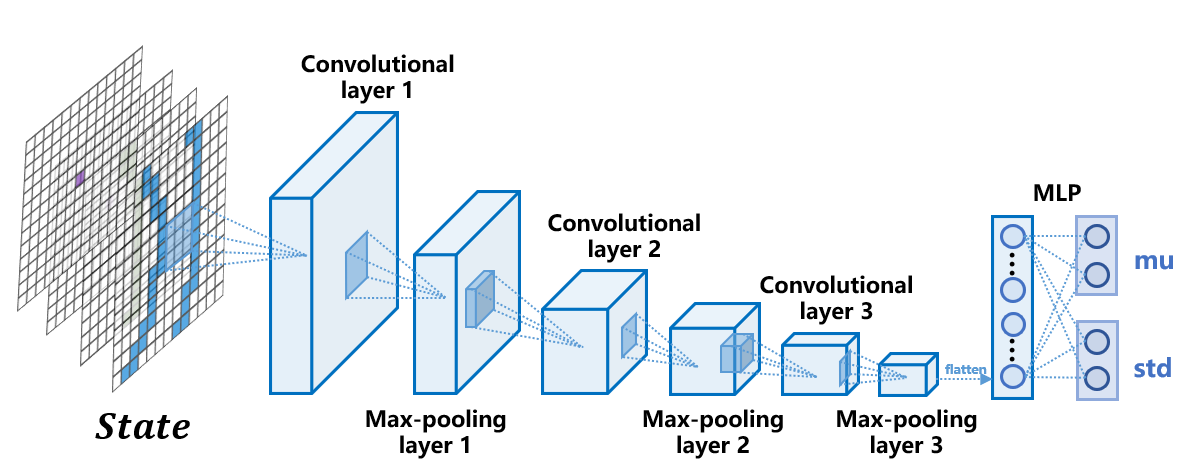}
        \caption{The Actor network architecture for action distribution output.}
        \label{fig:actor_net}
    \end{subfigure}
    
    \begin{subfigure}{\columnwidth}
        \centering
        \includegraphics[width=0.9\linewidth]{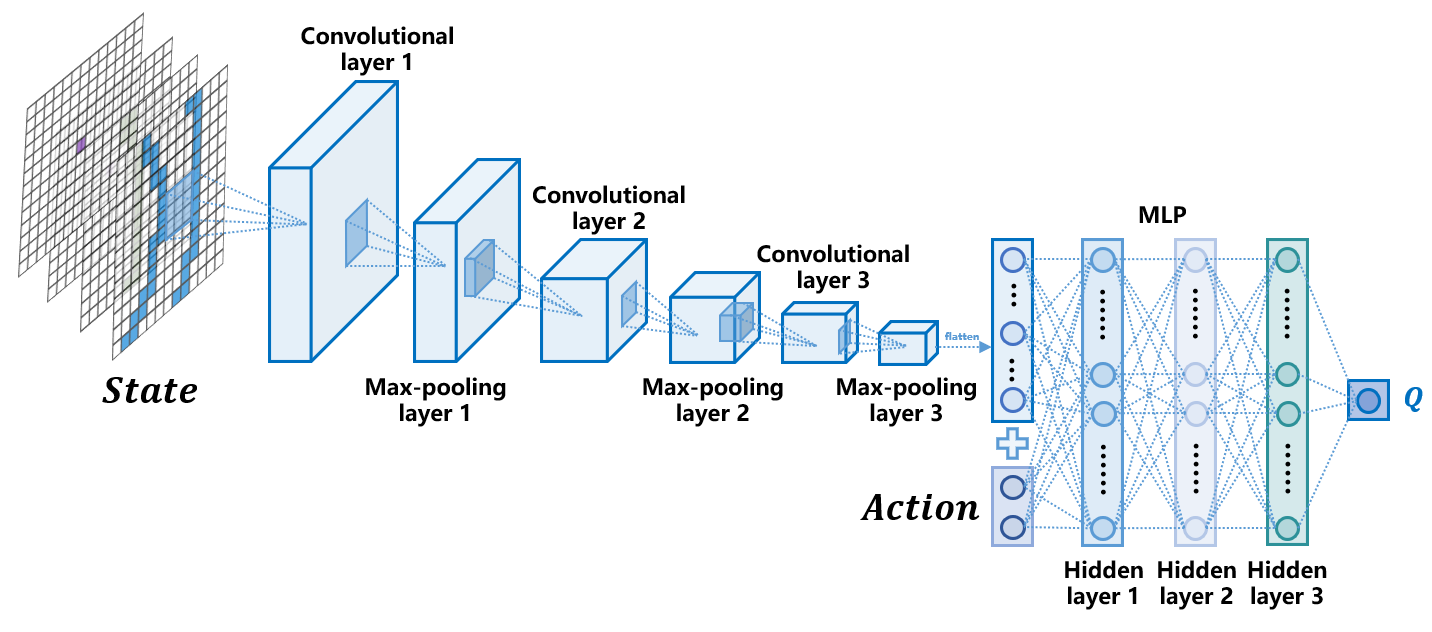}
        \caption{The Critic (and Cost-Critic) network architecture for value estimation.}
        \label{fig:critic_net}
    \end{subfigure}
    
    \caption{Schematic diagram of the proposed network architectures.}
    \label{fig:networks}
\end{figure}

\begin{itemize}[leftmargin=*, nosep]
    \item Training Scale: The total number of training steps is set to $T_{\max} = 250,000$. The agent begins the learning process after an initial exploration phase of $T_{\text{start}} = 20,000$ steps. The two-stage training step threshold is set to $T_{switch} = 200000$. The sliding-window length for stability cost smoothing is set to 40 steps. For performance evaluation, the number of test episodes is $K = 20$.
    \item Buffer Capacity: The capacity for both the replay buffer $\mathcal{D}$ and the cost replay buffer $\mathcal{D}_{\text{cost}}$ is set to $200,000$.
    \item Observation Setting: The ego-centric occupancy grid is constructed in the vehicle-aligned coordinate system. 
    The longitudinal and lateral ranges are set to \([-10,30]\) m and \([-7,7]\) m, respectively, with a grid resolution of 
    \(1\,\mathrm{m}\times1\,\mathrm{m}\). This results in a grid size of \(40\times14\). Since six feature channels are used in the observation, the CNN input tensor has a dimension of \(6\times40\times14\).
    \item Network Structure: The Actor network, shown in Fig.~\ref{fig:actor_net}, utilizes a convolutional neural network (CNN) backbone to extract features from the occupancy grid observation, followed by a Multi-Layer Perceptron (MLP) to output the mean and standard deviation of the action distribution. The critic and cost critic networks, shown in Fig.~\ref{fig:critic_net}, adopt a similar CNN structure for state processing. The extracted feature vectors are then concatenated with the action vector and fed into an MLP to estimate the Q-value of the state-action pair. The CNN backbone utilizes hidden channels of $[32, 64, 128]$. The subsequent Multi-Layer Perceptron (MLP) consists of hidden units with a configuration of $[256, 256, 256]$.
    \item Optimization: The learning rate for both the Actor and Critic networks is $1 \times 10^{-4}$, while the learning rate for the adaptive Lagrangian multipliers is set to $5 \times 10^{-5}$. The discount factor for future rewards and costs is $\gamma = 0.99$.
     \item Reward Setting: All reward components are first normalized and then combined into the total reward through a weighted summation. The speed thresholds are set to \([u_{h1},u_{h2}]=[10,70]~\mathrm{m/s}\) for the high-speed reward and \([u_{l1},u_{l2}]=[0,10]~\mathrm{m/s}\) for the low-speed penalty. The weights of different reward components are summarized in Table~\ref{tab:reward_weights}.
\end{itemize}

\begin{table}[htbp]
\centering
\caption{Reward weights used in the experiments.}
\label{tab:reward_weights}
\small
\setlength{\tabcolsep}{5pt}
\renewcommand{\arraystretch}{1.08}
\begin{tabular}{@{}lcc@{}}
\toprule
\textbf{Reward Component} & \textbf{Symbol} & \textbf{Weight} \\
\midrule
Track Constraint Reward       & $r_{\text{track}}$ & 0.2  \\
High-Speed Reward             & $r_{hs}$           & 0.6  \\
Low-Speed Reward              & $r_{ls}$           & -0.5 \\
Lap Completion Reward         & $r_{\text{lap}}$   & 0.1  \\
Trajectory Tracking Reward    & $r_{\text{MCRL}}$  & 0.6  \\
Target Speed Tracking Reward  & $r_{TS}$           & 0.6  \\
Heading Alignment Reward      & $r_{H}$            & 0.3  \\
\bottomrule
\end{tabular}
\end{table}

\begin{table}[htbp]
\centering
\caption{Vehicle dynamics parameters used in the experiment.}
\label{tab:vehicle_params}
\small
\setlength{\tabcolsep}{5pt}
\renewcommand{\arraystretch}{1.08}
\begin{tabular}{@{}lccc@{}}
\toprule
\textbf{Parameter} & \textbf{Symbol} & \textbf{Value} & \textbf{Unit} \\
\midrule
Mass                       & $m$                  & 800    & \unit{kg}     \\
Length                     & $L$                  & 4.7    & \unit{m}      \\
Width                      & $w_{\text{veh}}$     & 2      & \unit{m}      \\
CG to Front Axle           & $l_f$                & 2.115  & \unit{m}      \\
CG to Rear Axle            & $l_r$                & 1.88   & \unit{m}      \\
Front Cornering Stiffness  & $C_{\alpha f}$       & 64,000 & \unit{N/rad}  \\
Rear Cornering Stiffness   & $C_{\alpha r}$       & 64,000 & \unit{N/rad}  \\
\bottomrule
\end{tabular}
\end{table}

\textbf{Vehicle Setup} The vehicle dynamics parameters used in the experiment are summarized in \Cref{tab:vehicle_params}.

\subsection{Comparison Models}

\begin{itemize}[leftmargin=*, nosep]
    \item \textbf{Model Predictive Control (MPC):} A rule-based traditional control baseline commonly used in autonomous racing. 
    \item \textbf{Proximal Policy Optimization (PPO):} A robust on-policy algorithm that utilizes clipping for policy updates to ensure training stability.
    \item \textbf{Deep Deterministic Policy Gradient (DDPG):} A classic off-policy algorithm based on the deterministic policy gradient theorem.
    \item \textbf{Trajectory-Aided Learning (TAL)\cite{evans2023high}:}  A DRL framework that incorporates the optimal trajectory into the learning formulation to guide agents for high-speed racing.
     \item \textbf{Lagrangian-based Soft Actor Critic(SAC-Lag):} A safe RL baseline that incorporates explicit dynamics constraints into SAC through adaptive Lagrangian multipliers, where yaw-rate and sideslip-angle violations outside the safe operating envelope are directly penalized as safety costs.
    \item \textbf{TraD-RL:} The proposed dynamics-constrained and MCRL-guided reinforcement learning algorithm.
\end{itemize}

\subsection{Evaluating Metrics}

To evaluate the racing performance and safety level of the autonomous racing agents, the following performance metrics are employed:

\subsubsection{Racing Performance Metrics}

\begin{itemize}[leftmargin=*, nosep]
    \item \textbf{Lap Average Speed (LAS):} The average speed of the racing vehicle during a single lap. As a fundamental evaluation metric, it directly reflects the absolute racing capability of the agent on the track.
    \item \textbf{Lap Time (LT):} The time required for the racing vehicle to complete a single lap. By comprehensively considering both the driving speed and trajectory length, this metric effectively reflects the agent's global speed planning level and trajectory optimization capability.
\end{itemize}

\subsubsection{Safety Metrics}
\begin{itemize}[leftmargin=*, nosep]
    \item \textbf{Time-averaged Lap $\omega$-unsafe Times ($\omega$-TaUT):} The number of times the vehicle violates the yaw rate dynamic boundaries (i.e., boundaries \circled{1} and \circled{3} in Fig.~\ref{fig:safe_envelope}) within a single lap. To eliminate evaluation bias caused by varying lap completion times across different algorithms, this metric is normalized using the lap time.
    
    \item \textbf{Time-averaged Lap $\beta$-unsafe Times ($\beta$-TaUT):} The number of times the vehicle violates the center-of-mass sideslip angle dynamic boundaries (i.e., boundaries \circled{2} and \circled{4} in Fig.~\ref{fig:safe_envelope}) within a single lap. Similarly normalized by the lap time, this metric, together with the yaw rate violation times, constitutes a core indicator for evaluating the vehicle's dynamic stability at handling limits.
    
    \item \textbf{Lap Progress (LP):} The percentage of the track completed before the vehicle collides or runs off the track; if the lap is successfully completed, the progress reaches 100\%. This metric is utilized to macroscopically measure the overall safety and stability of the driving policy.
\end{itemize}

\subsection{Comparison Results and Analysis}

\begin{table*}[htbp]
    \centering
    \caption{Quantitative comparison of racing performance and safety metrics across different algorithms on different racetracks during the testing process. Values are presented as mean \(\pm\) standard deviation. Bold values denote the best results.}
    \label{tab:results_testing}
    \small
    \setlength{\tabcolsep}{4pt}
    \begin{tabular}{lcccccccc}
        \toprule
        & \multicolumn{4}{c}{Berlin Racetrack} 
        & \multicolumn{4}{c}{Modena Racetrack} \\
        \cmidrule(lr){2-5} \cmidrule(lr){6-9}
        & \multicolumn{2}{c}{Racing Performance} 
        & \multicolumn{2}{c}{Safety Performance}
        & \multicolumn{2}{c}{Racing Performance} 
        & \multicolumn{2}{c}{Safety Performance} \\
        \cmidrule(lr){2-3} \cmidrule(lr){4-5}
        \cmidrule(lr){6-7} \cmidrule(lr){8-9}
        Algorithm 
        & LAS (m/s) & LT (s) 
        & $\omega$-TaUT & $\beta$-TaUT
        & LAS (m/s) & LT (s)
        & $\omega$-TaUT & $\beta$-TaUT \\
        \midrule
        MPC 
        & $36.38 \pm 0.23$ & $64.24 \pm 0.41$ 
        & $8.98 \pm 0.34$ & $1.93 \pm 0.12$
        & $36.56 \pm 0.27$ & $56.39 \pm 0.44$
        & $\mathbf{10.54 \pm 0.54}$ & $\mathbf{1.72 \pm 0.07}$ \\
        
        DDPG 
        & $35.21 \pm 6.11$ & $63.61 \pm 11.09$ 
        & $18.66 \pm 0.18$ & $9.04 \pm 0.42$
        & $43.26 \pm 3.18$ & $47.51 \pm 4.46$
        & $19.37 \pm 0.37$ & $9.52 \pm 0.73$ \\
        
        PPO 
        & $30.23 \pm 0.64$ & $77.52 \pm 1.68$ 
        & $16.73 \pm 2.87$ & $3.65 \pm 0.52$
        & $36.59 \pm 3.29$ & $53.72 \pm 6.71$
        & $17.14 \pm 1.28$ & $4.61 \pm 0.34$ \\
        
        TAL 
        & $36.95 \pm 4.71$ & $62.41 \pm 16.86$ 
        & $18.18 \pm 2.09$ & $6.13 \pm 0.88$
        & $42.12 \pm 3.27$ & $46.51 \pm 3.77$
        & $19.07 \pm 0.64$ & $8.16 \pm 0.81$ \\
        
        SAC-Lag
        & $23.07 \pm 3.34$ & $104.63 \pm 14.57$ 
        & $\mathbf{7.74 \pm 1.29}$ & $\mathbf{0.91 \pm 0.57}$
        & $41.03 \pm 1.95$ & $50.10 \pm 2.51$
        & $16.45 \pm 1.12$ & $4.08 \pm 0.76$ \\
        
        TraD-RL 
        & $\mathbf{43.53 \pm 2.19}$ & $\mathbf{53.94 \pm 3.14}$ 
        & $17.39 \pm 0.68$ & $5.13 \pm 0.71$
        & $\mathbf{44.59 \pm 2.21}$ & $\mathbf{45.47 \pm 2.38}$
        & $17.12 \pm 1.20$ & $5.68 \pm 0.95$ \\
        \bottomrule
    \end{tabular}
\end{table*}

\subsubsection{Training Results}

\begin{figure*}[htbp]
    \centering
    \includegraphics[width=\textwidth]{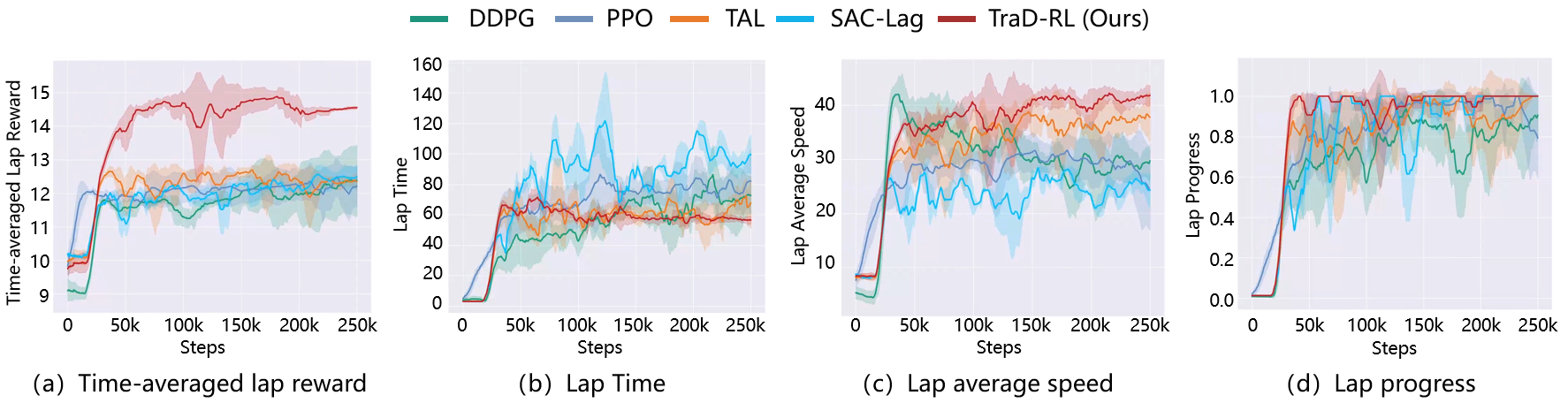}

    \caption{Learning curves of multi-dimensional performance metrics for different algorithms during the training process: 
    (a) time-averaged lap reward; (b) lap time; (c) lap average speed; and (d) lap progress. Solid lines denote the mean values, while the shaded regions represent 95\% confidence intervals over five runs.}
    \label{fig:training_curves}
\end{figure*}

Fig.~\ref{fig:training_curves}(a) illustrates the time-averaged lap reward (i.e., the ratio of the total cumulative lap reward to the completion time) of each algorithm during the training process, showing that all algorithms successfully reach a stable convergence state in the later stages of training. Meanwhile, Fig.~\ref{fig:training_curves}(b) and Fig.~\ref{fig:training_curves}(c) present the learning curves for lap time and lap average speed, respectively. As core metrics for evaluating the racing performance of autonomous vehicles, these two indicators intuitively reflect the agent's global speed planning level and comprehensive racing performance on complex tracks. Benefiting from the prior guidance of MCRL, the proposed method (Ours) stably achieves a lower lap time and a higher average lap speed after convergence compared to the other baseline algorithms, which thoroughly demonstrates that the incorporation of MCRL effectively breaks through the vehicle's racing bottlenecks. Furthermore, Fig.~\ref{fig:training_curves}(d) demonstrates the lap progress during training which indicates policy safety. Due to the complete absence of safety constraint mechanisms, the lap progress of the DDPG algorithm exhibits severe fluctuations and remains generally low; the policy clipping mechanism of PPO implicitly enhances training stability to a certain extent, resulting in slightly better completion performance than DDPG. Guided globally by the racing line, the TAL method effectively prevents the vehicle from deviating from the normal driving trajectory. However, due to the lack of explicit vehicle dynamics constraints, it still struggles to avoid safety violations under limit-handling conditions, such as high-speed cornering. In contrast, by synergistically integrating the MCRL global guidance and explicit dynamics constraints, the proposed method stably ensures a 100\% lap progress after approximately 15k training steps, demonstrating remarkable safety improvement and training stability in high-speed racing scenarios.

\subsubsection{Testing Results}

\begin{figure*}[htbp]
    \centering

    \begin{subfigure}{0.24\textwidth}
        \centering
        \includegraphics[width=\linewidth]{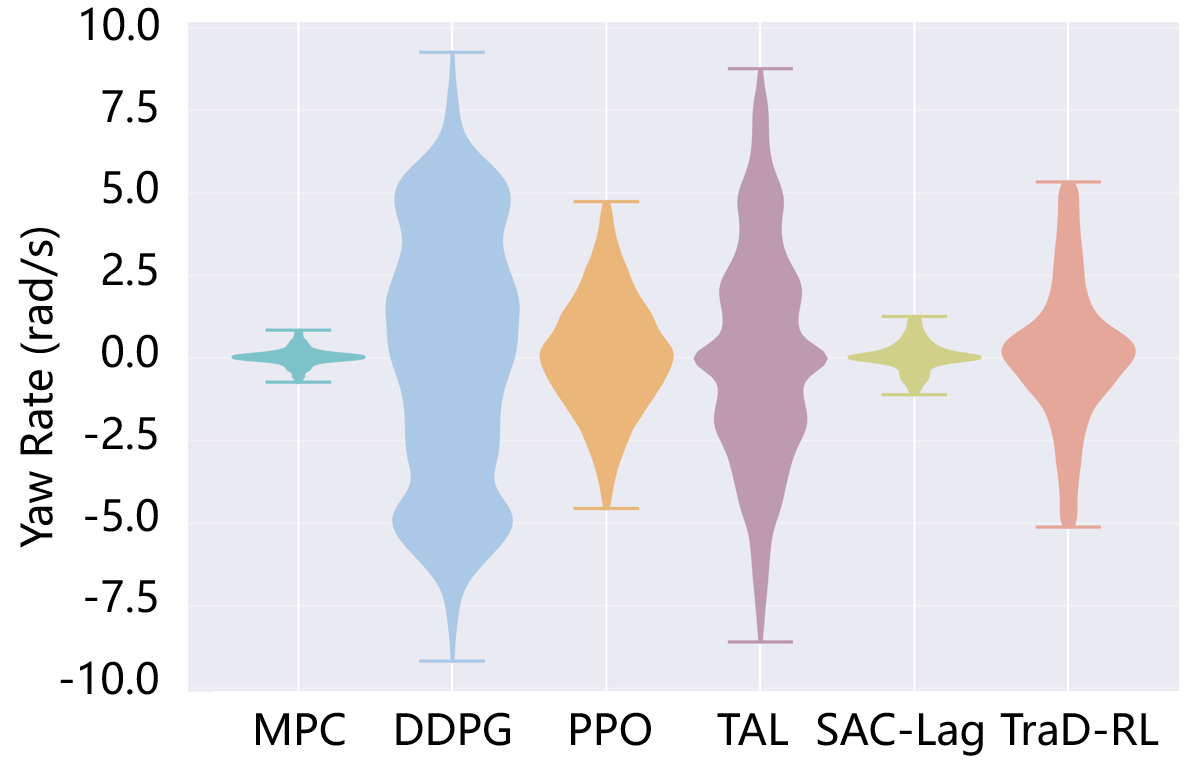}
        \caption{Yaw rate}
        \label{fig:test_violin_yaw}
    \end{subfigure}
    \hfill
    \begin{subfigure}{0.245\textwidth}
        \centering
        \includegraphics[width=\linewidth]{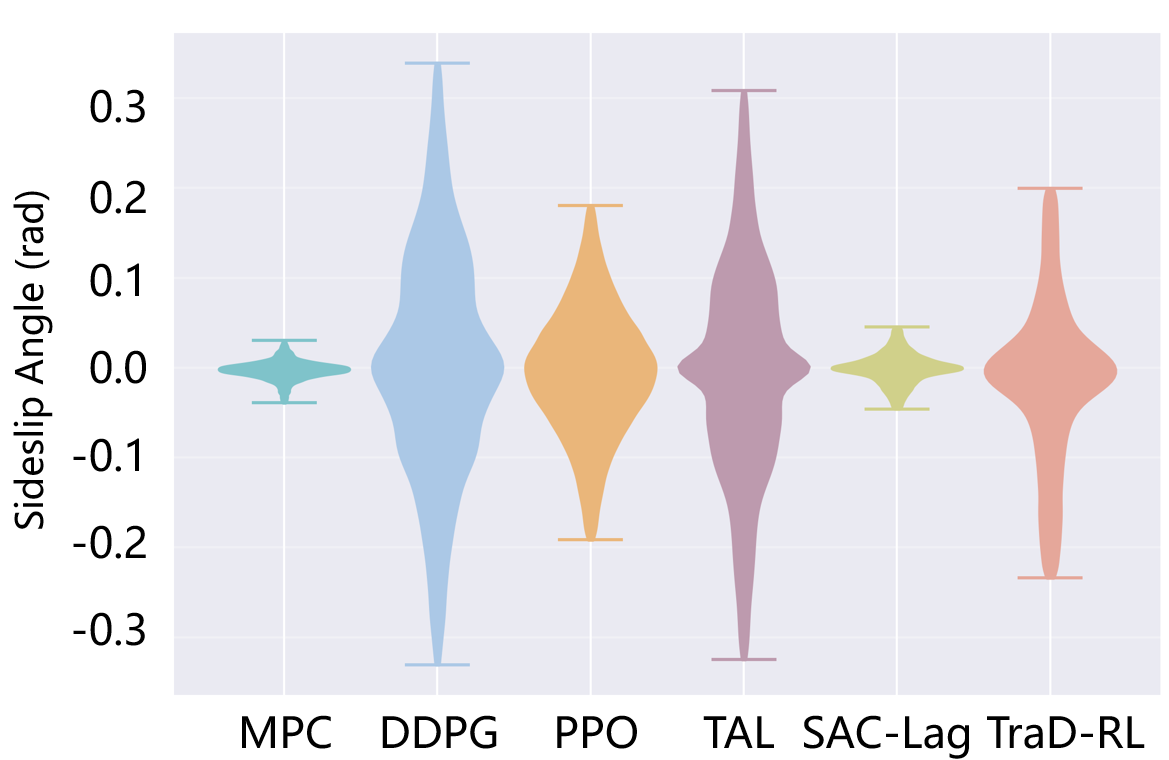}
        \caption{Sideslip angle}
        \label{fig:test_violin_beta}
    \end{subfigure}
    \hfill
    \begin{subfigure}{0.245\textwidth}
        \centering
        \includegraphics[width=\linewidth]{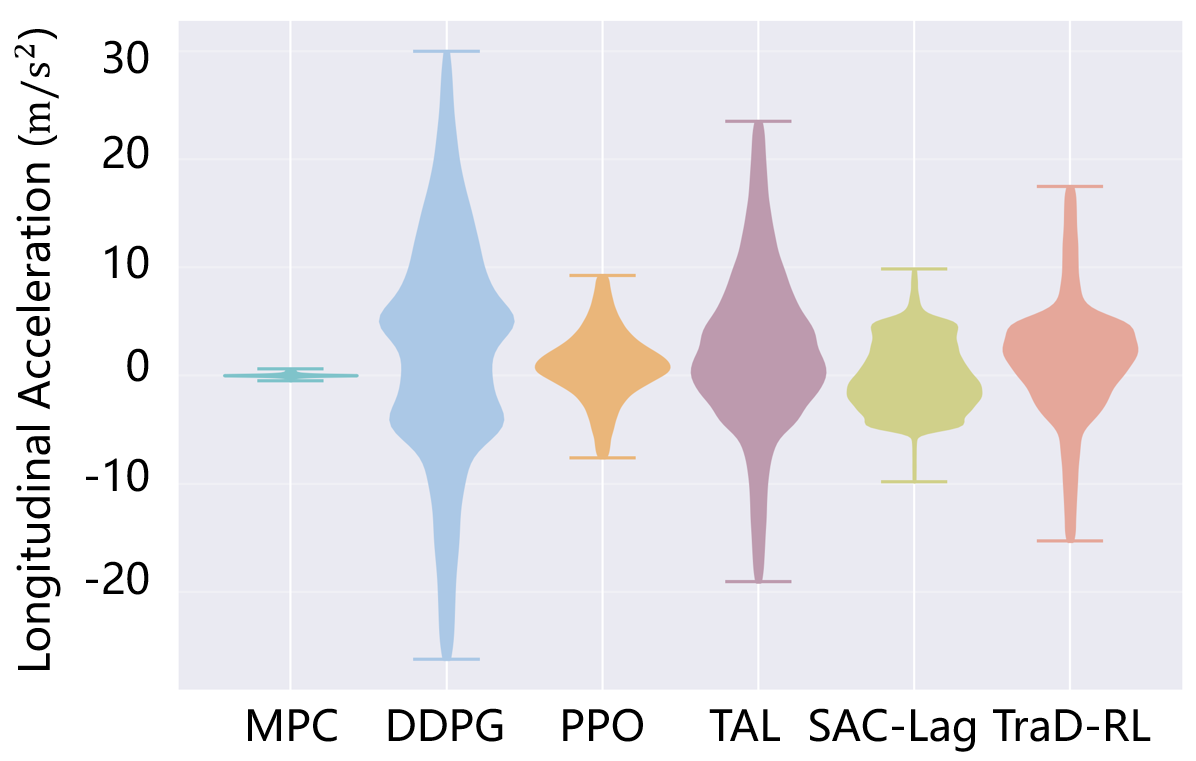}
        \caption{Longitudinal acceleration}
        \label{fig:test_violin_acc}
    \end{subfigure}
    \hfill
    \begin{subfigure}{0.24\textwidth}
        \centering
        \includegraphics[width=\linewidth]{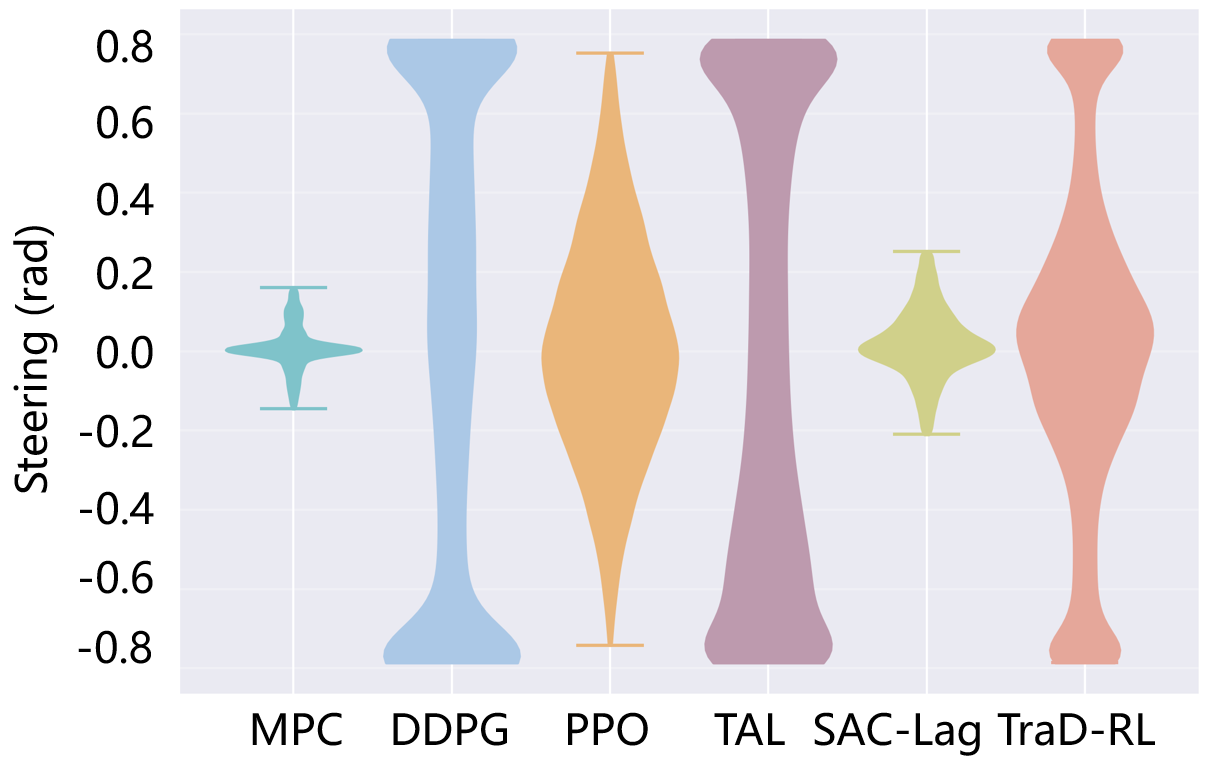}
        \caption{Steering angle}
        \label{fig:test_violin_steer}
    \end{subfigure}
    
    \caption{Statistical distributions of control inputs and vehicle dynamic states for different algorithms during testing on the Berlin Racetrack: 
    (a) yaw rate; (b) sideslip angle; (c) longitudinal acceleration; and (d) steering angle.}
    \label{fig:test_violin_plots_berlin}
\end{figure*}

The racing performance metrics in \Cref{tab:results_testing} show that the proposed TraD-RL method achieves the best lap performance on both the Berlin and Modena racetracks, demonstrating good cross-track generalization capability. On the Berlin Racetrack, TraD-RL improves the lap average speed by 19.65\%, 23.63\%, 43.99\%, 17.81\%, and 88.69\% over MPC, DDPG, PPO, TAL, and SAC-Lag, respectively; the corresponding lap time is reduced by 16.03\%, 15.20\%, 30.42\%, 13.57\%, and 48.45\%, respectively. On the Modena Racetrack, TraD-RL also achieves the highest lap average speed and the shortest lap time. These results indicate that the path reference and velocity guidance introduced by MCRL can effectively improve the agent's track utilization and high-speed cornering capability, enabling superior racing performance under different track geometries.

Regarding the safety metrics, \Cref{tab:results_testing} shows that TraD-RL effectively mitigates dynamic instability while improving racing performance. Compared with DDPG and TAL, TraD-RL consistently reduces both $\omega$-TaUT and $\beta$-TaUT on the Berlin and Modena Racetracks. In particular, $\beta$-TaUT is reduced by 43.25\% and 16.31\% on the Berlin Racetrack, and by 40.34\% and 30.39\% on the Modena Racetrack, respectively. Although MPC and SAC-Lag achieve lower unsafe times in some cases, their lap average speed and lap time are worse, indicating more conservative driving behaviors. In particular, MPC provides stable reference-tracking performance, but its conservative optimization-based control makes it difficult to further exploit the upper performance limit of high-speed racing. In contrast, the CBF-inspired dynamics constraints in TraD-RL guide the policy to operate near the boundary of the safe operating envelope while avoiding severe instability. Therefore, TraD-RL achieves a better performance--safety trade-off for high-speed autonomous racing.

\begin{figure*}[htbp]
    \centering

    \begin{subfigure}{0.24\textwidth}
        \centering
        \includegraphics[width=\linewidth]{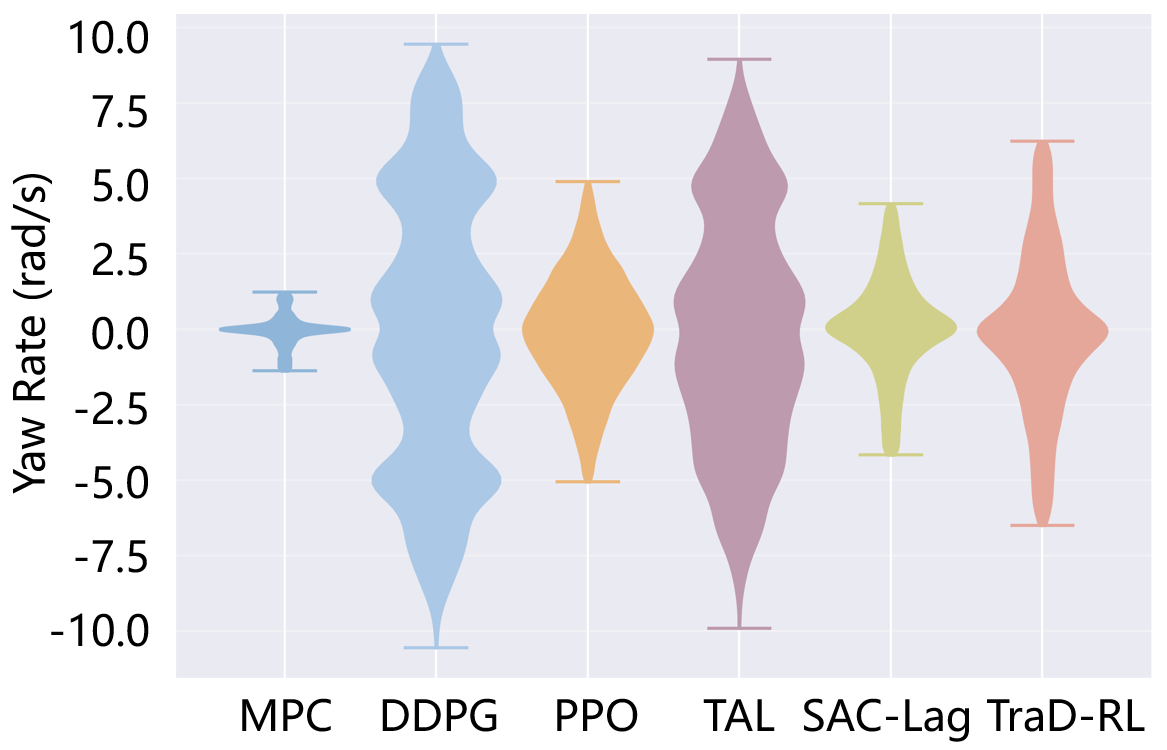}
        \caption{Yaw rate}
        \label{fig:test_violin_yaw_modena}
    \end{subfigure}
    \hfill
    \begin{subfigure}{0.247\textwidth}
        \centering
        \includegraphics[width=\linewidth]{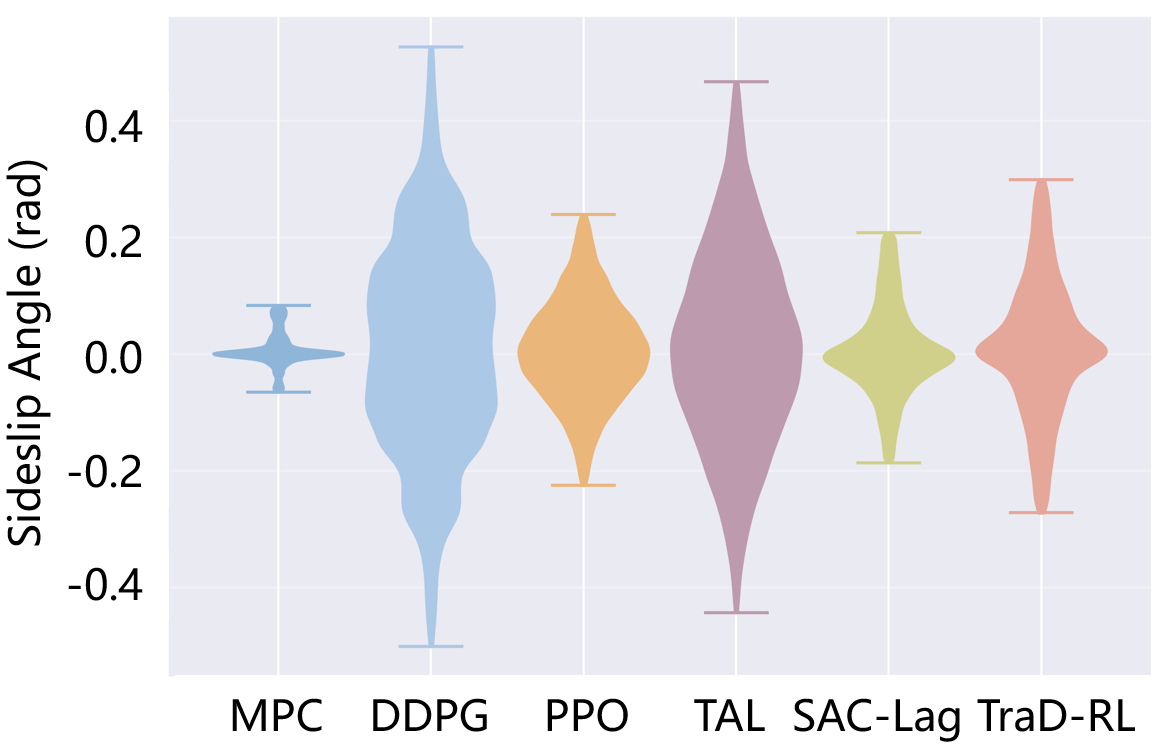}
        \caption{Sideslip angle}
        \label{fig:test_violin_beta_modena}
    \end{subfigure}
    \hfill
    \begin{subfigure}{0.245\textwidth}
        \centering
        \includegraphics[width=\linewidth]{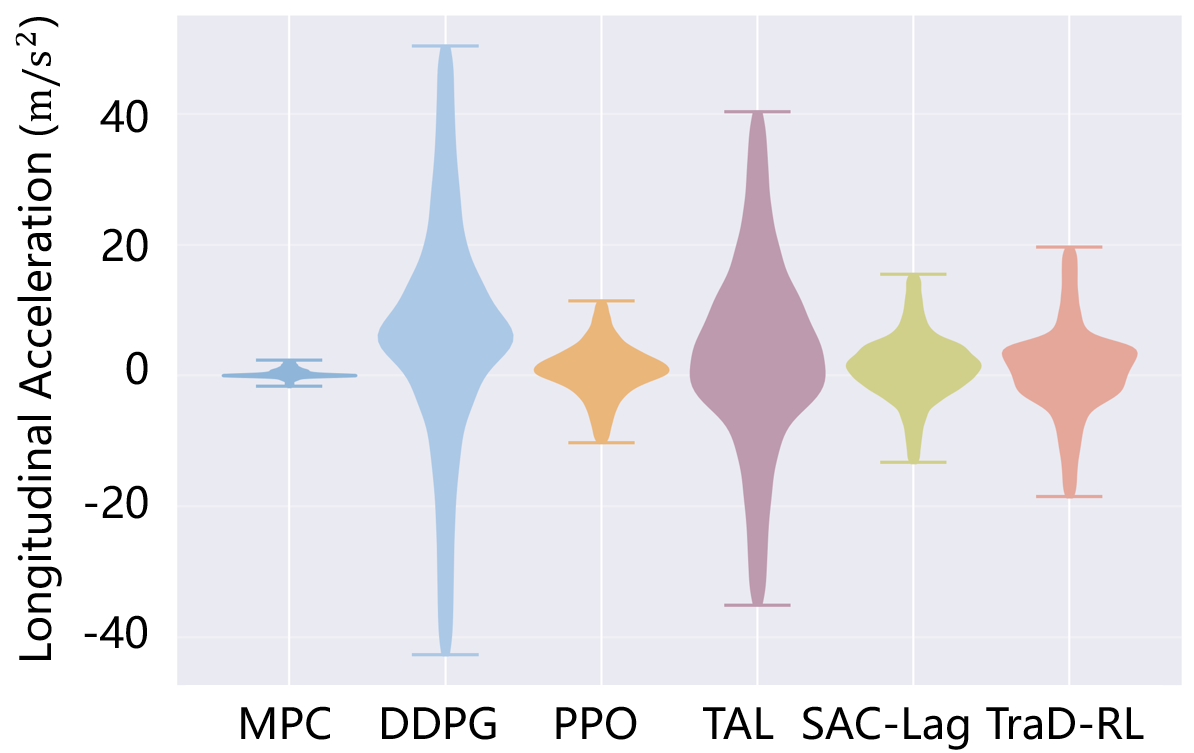}
        \caption{Longitudinal acceleration}
        \label{fig:test_violin_acc_modena}
    \end{subfigure}
    \hfill
    \begin{subfigure}{0.24\textwidth}
        \centering
        \includegraphics[width=\linewidth]{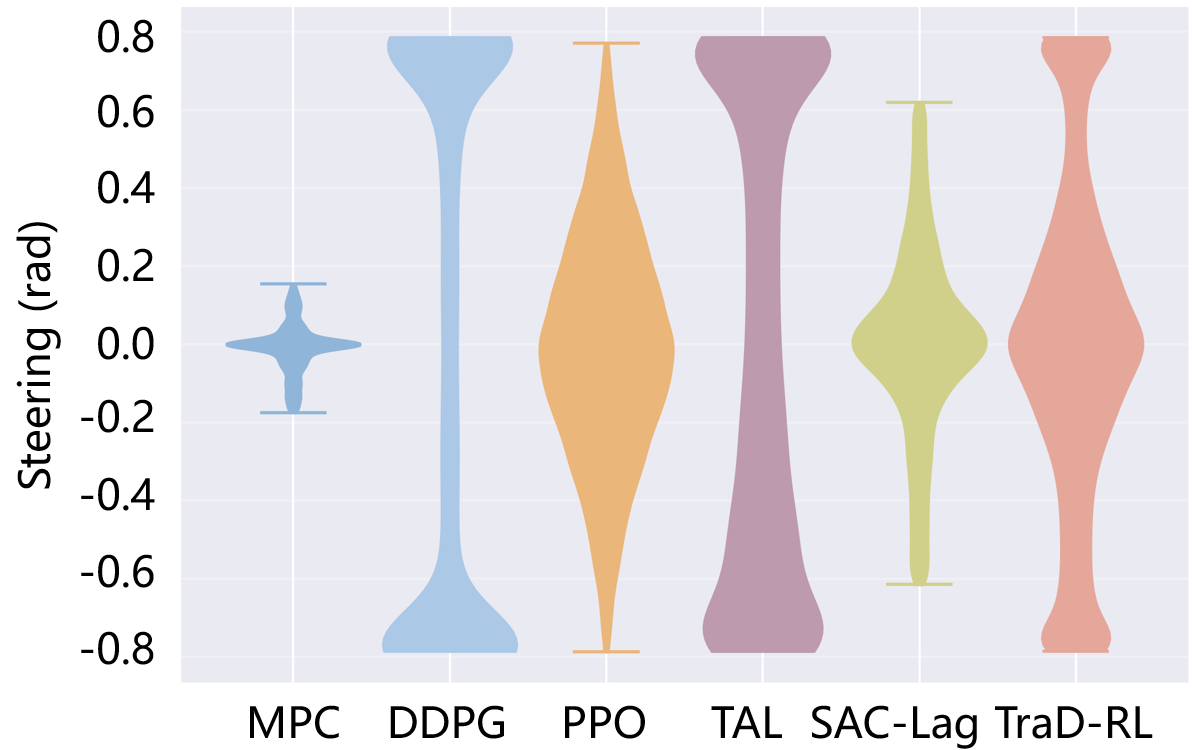}
        \caption{Steering angle}
        \label{fig:test_violin_steer_modena}
    \end{subfigure}
    
    \caption{Statistical distributions of control inputs and vehicle dynamic states for different algorithms during testing on the Modena Racetrack: 
    (a) yaw rate; (b) sideslip angle; (c) longitudinal acceleration; and (d) steering angle.}
    \label{fig:test_violin_plots_modena}
\end{figure*}

\Cref{fig:test_violin_plots_berlin} and \Cref{fig:test_violin_plots_modena} show the distributions of control inputs and vehicle dynamics states during testing on the Berlin and Modena racetracks. DDPG and TAL exhibit wider distributions in longitudinal acceleration, steering angle, yaw rate, and sideslip angle, with more pronounced long-tail patterns. This indicates larger control fluctuations and a higher risk of dynamic instability, and also reflects the action oscillation problem commonly observed in unconstrained or weakly constrained reinforcement learning policies. In contrast, MPC, PPO, and SAC-Lag show more concentrated distributions. For MPC, the distributions of yaw rate, sideslip angle, and control inputs are noticeably narrower than those of most reinforcement learning methods, indicating that optimization-based control can provide smoother and more stable outputs in reference-tracking tasks. By contrast, some reinforcement learning policies directly output actions through neural networks and are therefore more prone to control oscillations and state fluctuations. However, the concentrated action distribution of MPC also suggests relatively conservative control behavior, making it difficult to further exploit the performance upper bound in high-speed racing tasks. For SAC-Lag, the steering angle and vehicle states are mostly distributed around zero, reflecting a conservative driving strategy. This helps reduce dynamic risks, but also limits the utilization of track width and tire adhesion limits. Compared with these methods, TraD-RL shows more balanced distribution characteristics on both racetracks. Its yaw rate and sideslip angle distributions are more concentrated than those of DDPG and TAL, with fewer extreme values. At the same time, its longitudinal acceleration and steering angle still retain a reasonable action range. These results show that TraD-RL can suppress unnecessary dynamic fluctuations while maintaining high racing performance, demonstrating a better balance between racing performance and vehicle dynamic stability.

\begin{figure*}[htbp] 
    \centering
    \includegraphics[width=0.8\textwidth]{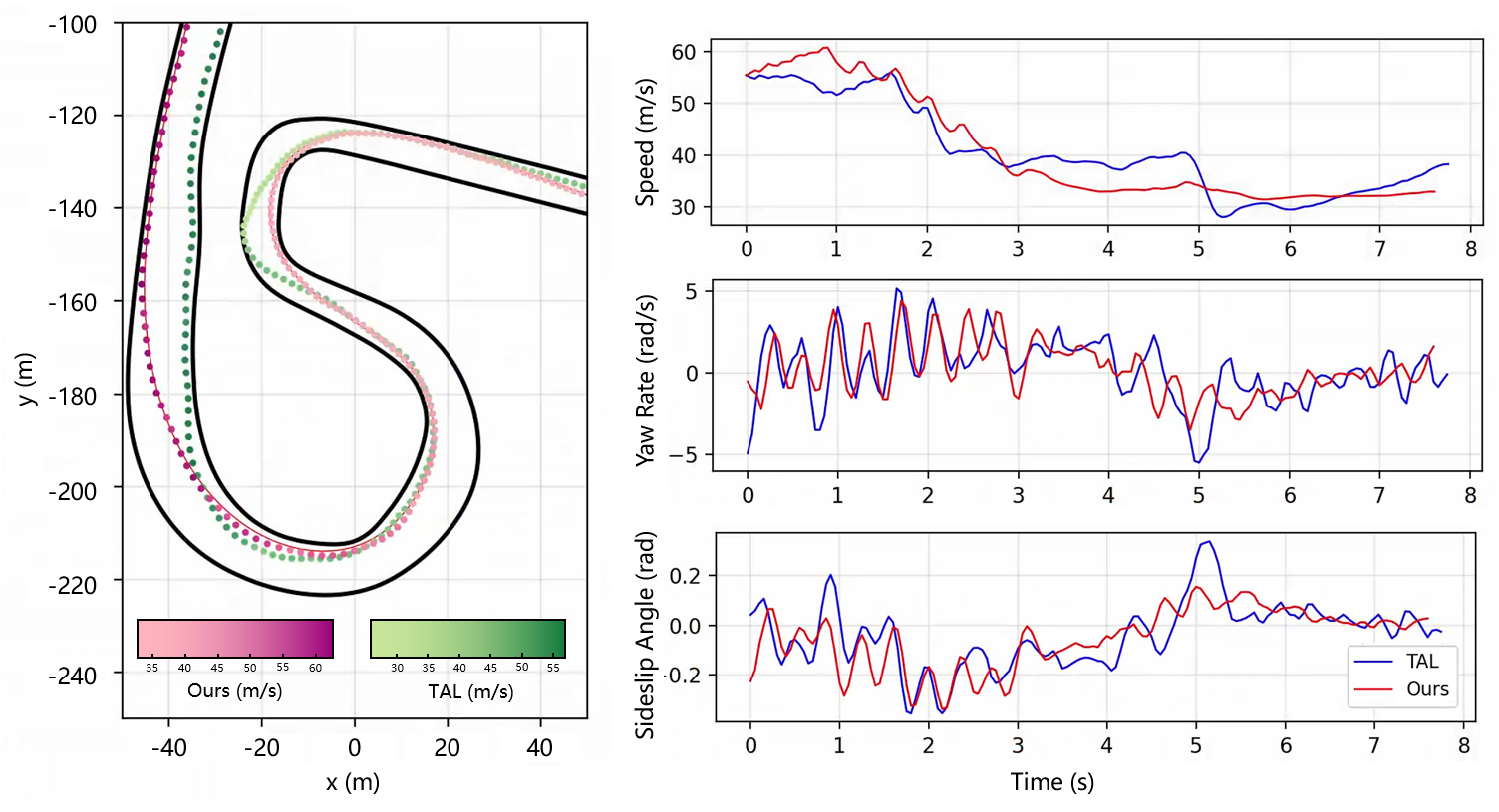} 
    \caption{Experimental results comparison in a continuous corner (S-curve) section of the Berlin Tempelhof Airport Street Circuit. Left: Trajectory and speed heat map distributions of TAL and the proposed method (Ours), with the red solid line representing the MCRL. Right: Time-series comparison of vehicle speed, yaw rate, and sideslip angle during the cornering process.}
    \label{fig:case study}
\end{figure*}
\begin{figure}
    \centering
    \includegraphics[width=0.85\columnwidth]{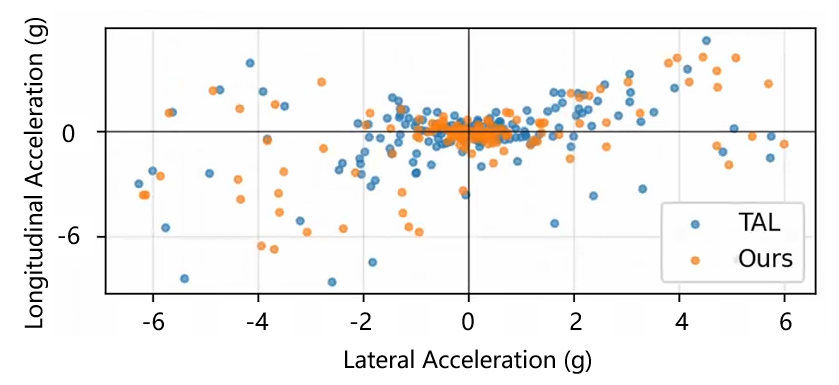}
    \caption{G-G diagram comparison of longitudinal and lateral acceleration distributions in the continuous corner (S-curve).}
    \label{fig:case_study_gg}
\end{figure}

\subsubsection{Case Study}

Fig. \ref{fig:case study} (left) visualizes the trajectory and speed distributions of the two methods in a continuous corner. It can be observed that TAL exhibits evident trajectory discontinuities and local oscillations during corner entry and exit, resulting in zigzag-like trajectory deviations. In contrast, guided by the global racing-line prior provided by \(o_{\text{MCRL}}\), TraD-RL generates a smoother and more continuous driving trajectory, with more natural curvature transitions at the corner entry, mid-corner, and corner exit. This indicates that MCRL guidance can provide effective global path and velocity references for the policy, thereby improving trajectory consistency and control coherence in continuous cornering. According to the velocity profile (Fig. \ref{fig:case study}, top-right), TraD-RL maintains a higher speed before corner entry and completes corner entry through a smoother deceleration process. In comparison, TAL exhibits a clear speed drop in the middle-to-late part of the corner, indicating insufficient coordination between trajectory tracking and velocity control in continuous cornering. When the vehicle approaches the track boundary or experiences rapid attitude changes, the policy has to apply stronger braking to correct the vehicle state, resulting in additional speed loss. Therefore, TraD-RL can pass through the continuous corner with more coherent trajectory and velocity control, achieving higher local driving efficiency.

Furthermore, \Cref{fig:case_study_gg} shows the distribution of longitudinal and lateral accelerations in this case study. The wide distribution range of both longitudinal and lateral accelerations in the G-G diagram indicates that the vehicle enters a high-dynamic nonlinear handling condition in the continuous corner, where longitudinal acceleration/deceleration and lateral cornering demands are strongly coupled. This imposes higher requirements on yaw stability and sideslip stability control. As shown by the yaw rate and sideslip angle curves (Fig. \ref{fig:case study}, middle-right and bottom-right), TAL exhibits stronger state oscillations and larger peak values during cornering, indicating that the vehicle is more likely to approach unstable dynamic states. In contrast, TraD-RL produces smoother yaw-rate and sideslip-angle responses with clearly reduced peak amplitudes, demonstrating that the explicit dynamics constraints can effectively mitigate excessive yaw response and sideslip fluctuations. Overall, this case study shows that TraD-RL not only improves trajectory coherence and velocity efficiency in continuous cornering, but also enhances vehicle dynamic stability under high-dynamic nonlinear handling conditions.

\subsection{Ablation Experiment}

\begin{table*}[htbp]
    \centering
    \caption{Quantitative comparison of racing performance and safety metrics across different ablation algorithms during the testing process. Values are presented as mean \(\pm\) standard deviation.}
    \label{tab:ablation_results}
    \small
    \setlength{\tabcolsep}{4pt}
    \begin{tabular}{lcccccccc}
        \toprule
        & \multicolumn{4}{c}{Berlin Racetrack} 
        & \multicolumn{4}{c}{Modena Racetrack} \\
        \cmidrule(lr){2-5} \cmidrule(lr){6-9}
        & \multicolumn{2}{c}{Racing Performance} 
        & \multicolumn{2}{c}{Safety Performance}
        & \multicolumn{2}{c}{Racing Performance} 
        & \multicolumn{2}{c}{Safety Performance} \\
        \cmidrule(lr){2-3} \cmidrule(lr){4-5}
        \cmidrule(lr){6-7} \cmidrule(lr){8-9}
        Method
        & LAS (m/s) & LT (s) 
        & $\omega$-TaUT & $\beta$-TaUT
        & LAS (m/s) & LT (s)
        & $\omega$-TaUT & $\beta$-TaUT \\
        \midrule
        w/o DC
        & $50.45 \pm 2.90$ & $46.05 \pm 2.98$
        & $19.23 \pm 0.44$ & $10.01 \pm 1.17$
        & $49.26 \pm 3.02$ & $41.14 \pm 3.23$
        & $19.23 \pm 0.95$ & $10.58 \pm 0.85$ \\
        
        w/o TG
        & $19.95 \pm 2.86$ & $115.28 \pm 16.56$
        & $12.76 \pm 0.55$ & $6.94 \pm 0.96$
        & $28.92 \pm 3.36$ & $73.80 \pm 13.76$
        & $16.50 \pm 0.88$ & $7.06 \pm 5.68$ \\
        
        Ours
        & $43.53 \pm 2.19$ & $53.94 \pm 3.14$
        & $17.39 \pm 0.68$ & $5.13 \pm 0.71$
        & $44.59 \pm 2.21$ & $45.47 \pm 2.38$
        & $17.12 \pm 1.20$ & $5.68 \pm 0.95$ \\
        \bottomrule
    \end{tabular}
\end{table*}

\begin{figure*}[htbp]
    \centering

    \begin{subfigure}{0.24\textwidth}
        \centering
        \includegraphics[width=\linewidth]{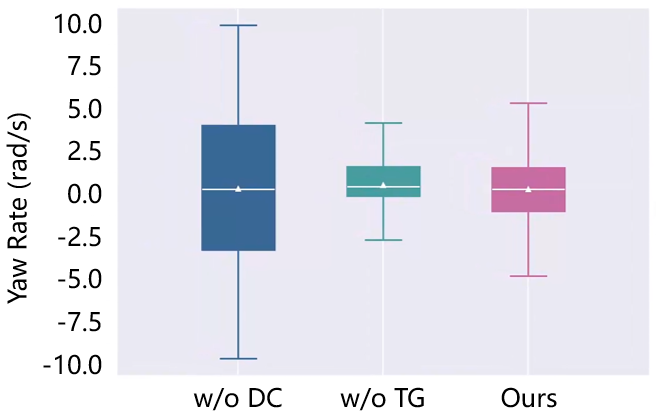}
        \caption{Yaw rate}
        \label{fig:abl_yaw}
    \end{subfigure}
    \hfill
    \begin{subfigure}{0.243\textwidth}
        \centering
        \includegraphics[width=\linewidth]{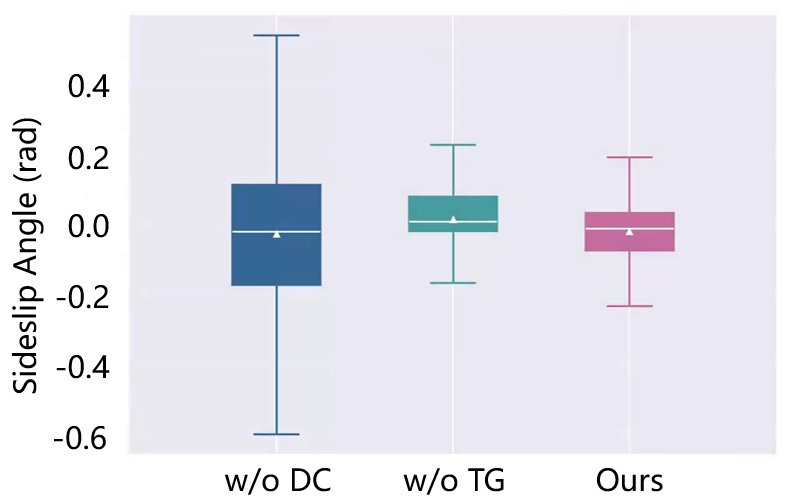}
        \caption{Sideslip angle}
        \label{fig:abl_beta}
    \end{subfigure}
    \hfill
    \begin{subfigure}{0.243\textwidth}
        \centering
        \includegraphics[width=\linewidth]{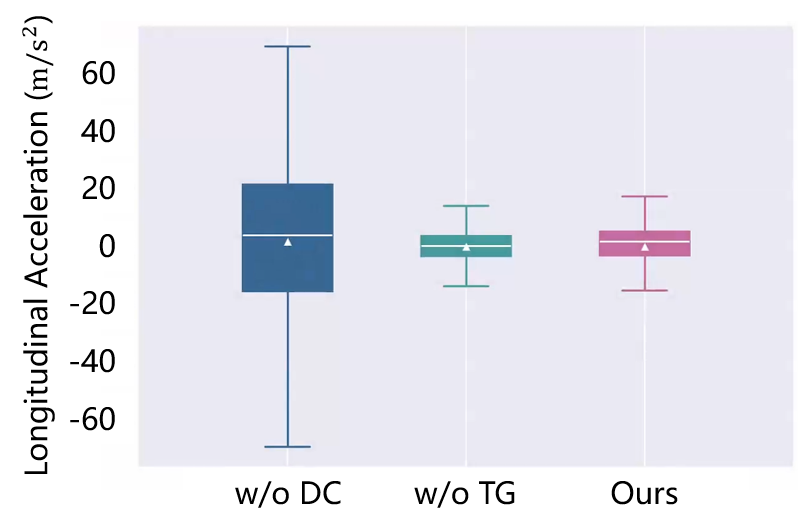}
        \caption{Longitudinal acceleration}
        \label{fig:abl_acc}
    \end{subfigure}
    \hfill
    \begin{subfigure}{0.24\textwidth}
        \centering
        \includegraphics[width=\linewidth]{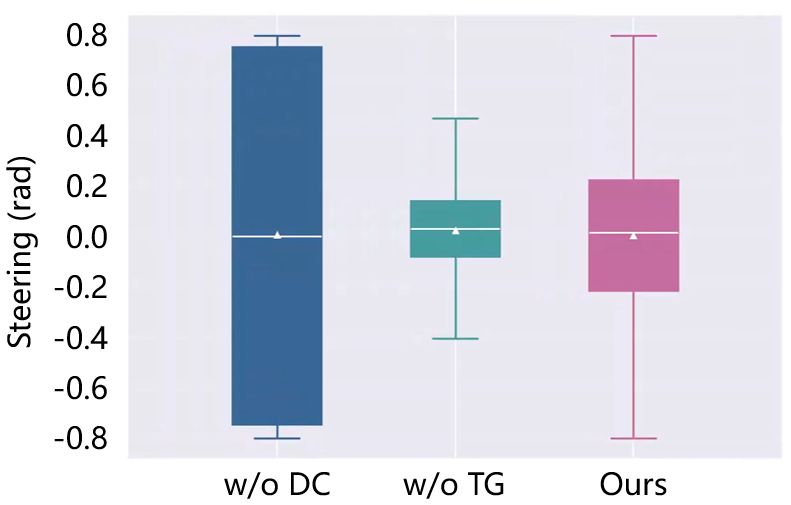}
        \caption{Steering angle}
        \label{fig:abl_steering}
    \end{subfigure}

    \caption{Control-output and dynamic-state comparison of different ablation algorithms during the testing phase: 
    (a) yaw rate; (b) sideslip angle; (c) longitudinal acceleration; and (d) steering angle.}
    \label{fig:ablation_analysis}
\end{figure*}

To further validate the effectiveness of the core modules in the proposed TraD-RL algorithm, we conducted ablation studies on both the Berlin and Modena racetracks by removing the trajectory guidance module (w/o TG) and the explicit dynamics constraints module (w/o DC). Table \ref{tab:ablation_results} reports the racing performance and stability metrics of different ablation algorithms on the two racetracks, while Fig. \ref{fig:ablation_analysis} illustrates the corresponding control-output and dynamic-state distributions.

As shown in Table \ref{tab:ablation_results}, removing the trajectory guidance module causes a significant degradation in racing performance. On the Berlin racetrack, w/o TG achieves an average lap speed of only \(19.95\) m/s and a lap time of \(115.28\) s. whereas the full TraD-RL model reaches \(43.53\) m/s and \(53.94\) s, corresponding to a \(118.20\%\) improvement in average speed and a \(53.21\%\) reduction in lap time. The results on the Modena racetrack show a similar trend, which demonstrates that MCRL trajectory guidance provides effective expert knowledge priors, thereby improving track utilization and speed performance.

The contribution of the explicit dynamics constraints is verified by comparing the full TraD-RL model with w/o DC. Although w/o DC achieves a higher average speed and shorter lap time on the Berlin racetrack, it also produces more dynamic constraint violations. In comparison, the full TraD-RL model reduces the time-averaged lap \(\omega\)-TaUT and \(\beta\)-TaUT by \(9.57\%\) and \(48.75\%\), respectively. This confirms that without explicit dynamics constraints, the agent tends to adopt more aggressive control actions, making the yaw rate and sideslip angle more likely to approach or exceed the stability boundaries. The results on the Modena racetrack further support this observation.

Fig. \ref{fig:ablation_analysis} further ompares the control-output and dynamic-state distributions of different ablation variants. As shown in Fig. \ref{fig:ablation_analysis}(\subref{fig:abl_acc}) and Fig. \ref{fig:ablation_analysis}(\subref{fig:abl_steering}), w/o DC produces much wider distributions of longitudinal acceleration and steering, indicating more aggressive and fluctuating control actions. This leads to broader yaw-rate and sideslip-angle distributions, as shown in Fig. \ref{fig:ablation_analysis}(\subref{fig:abl_yaw}) and Fig. \ref{fig:ablation_analysis}(\subref{fig:abl_beta}), , suggesting a higher risk of approaching unstable dynamic states. By introducing the explicit dynamics constraints, the proposed method can effectively suppress excessive control fluctuations and extreme dynamic responses. Moreover, although w/o TG shows compact distributions in some metrics, this is mainly caused by its low-speed conservative behavior rather than improved high-speed stability.


\subsection{Sensitivity and Robustness Analysis}

\begin{figure}
    \centering
    \includegraphics[width=0.6\columnwidth]{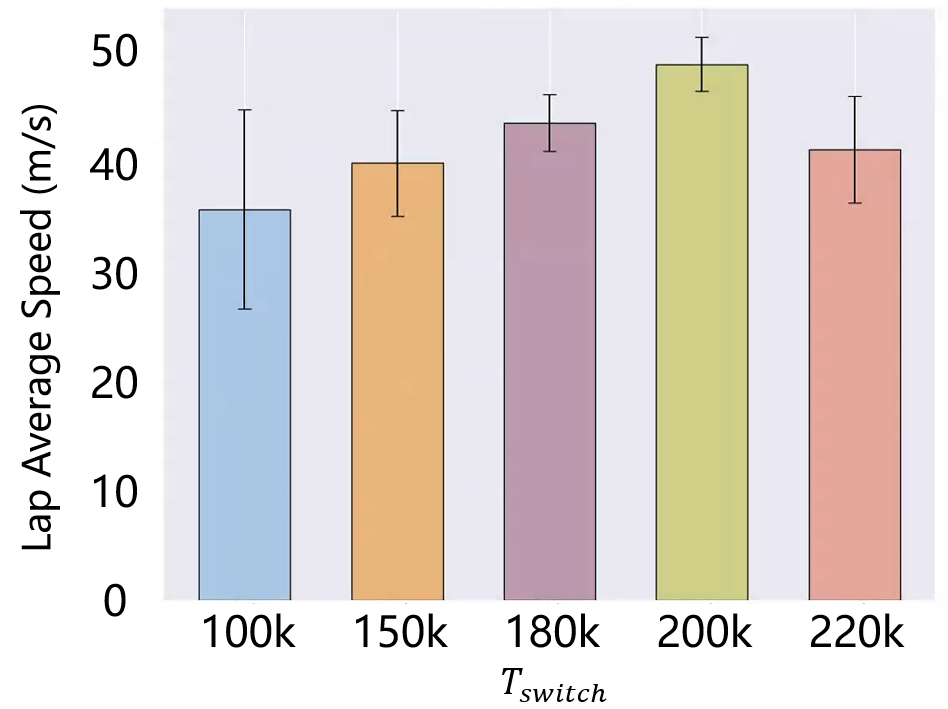}
    \caption{Sensitivity analysis of the curriculum switching step \(T_{\mathrm{switch}}\). The lap average speed is compared under different switching steps during the testing phase.}
    \label{fig:switch_sensitivity}
\end{figure}

We conducted a sensitivity analysis of \(T_{\mathrm{switch}}\) to analyze the influence of the curriculum switching step on training performance. Fig. \ref{fig:switch_sensitivity} presents the testing average lap speed under different \(T_{\mathrm{switch}}\) settings with a total training budget of 250k steps. It can be observed that early switching, such as \(T_{\mathrm{switch}}=100k\) or \(150k\), makes the agent enter the high-speed exploration stage before the trajectory-following and basic speed-control policies become sufficiently stable, thereby limiting the final racing performance. As \(T_{\mathrm{switch}}\) increases, the testing speed gradually improves and reaches the highest value at \(T_{\mathrm{switch}}=200k\). In contrast, when \(T_{\mathrm{switch}}=220k\), only 30k steps remain for high-speed exploration, which is insufficient for fully exploiting the vehicle's performance limits. Therefore, \(T_{\mathrm{switch}}=200k\) is adopted in this work to balance stable trajectory-guided learning and high-speed performance exploration.

To evaluate the robustness of the proposed method under observation disturbances, Gaussian noise with different intensities was further added to the observation inputs during testing. As shown in Table \ref{tab:noise_robustness}, the average lap speed gradually decreases as the noise level increases, while the lap time increases accordingly. This indicates that observation noise can affect the agent's perception of the track and vehicle states, leading to a certain degradation in racing performance. Nevertheless, the policy still maintains stable lap completion under both moderate and high noise conditions. Meanwhile, the dynamic-state violation counts do not deteriorate with increasing noise. This is mainly because the policy tends to become more conservative under noisy observations, reducing the frequency of entering extreme unstable states. Overall, TraD-RL does not exhibit obvious policy failure under different noise conditions, demonstrating its robustness to observation uncertainty.

\begin{table}[htbp]
\centering
\caption{Robustness analysis under different observation noise levels on the Berlin racetrack. Values are presented as mean \(\pm\) standard deviation.}
\label{tab:noise_robustness}
\small
\setlength{\tabcolsep}{3pt}
\renewcommand{\arraystretch}{1.25}
\resizebox{\columnwidth}{!}{
\begin{tabular}{lcccc}
\hline
\multirow{2}{*}{Noise Level} 
& \multicolumn{2}{c}{Racing Performance} 
& \multicolumn{2}{c}{Safety Performance} \\
\cline{2-5}
& LAS (m/s) & LT (s) & \(\omega\)-TaUT & \(\beta\)-TaUT \\
\hline
No Noise       & \(43.68 \pm 2.23\) & \(53.75 \pm 3.12\) & \(17.42 \pm 0.77\) & \(5.22 \pm 0.89\) \\
Moderate Noise & \(42.63 \pm 1.98\) & \(55.13 \pm 2.85\) & \(16.68 \pm 0.72\) & \(4.40 \pm 0.70\) \\
Strong Noise   & \(41.20 \pm 1.94\) & \(57.16 \pm 3.04\) & \(16.69 \pm 0.69\) & \(3.87 \pm 0.59\) \\
\hline
\end{tabular}
}
\end{table}

\subsection{Computational Efficiency Analysis}

\begin{figure}
    \centering

    \begin{subfigure}{0.49\columnwidth}
        \centering
        \includegraphics[width=\linewidth]{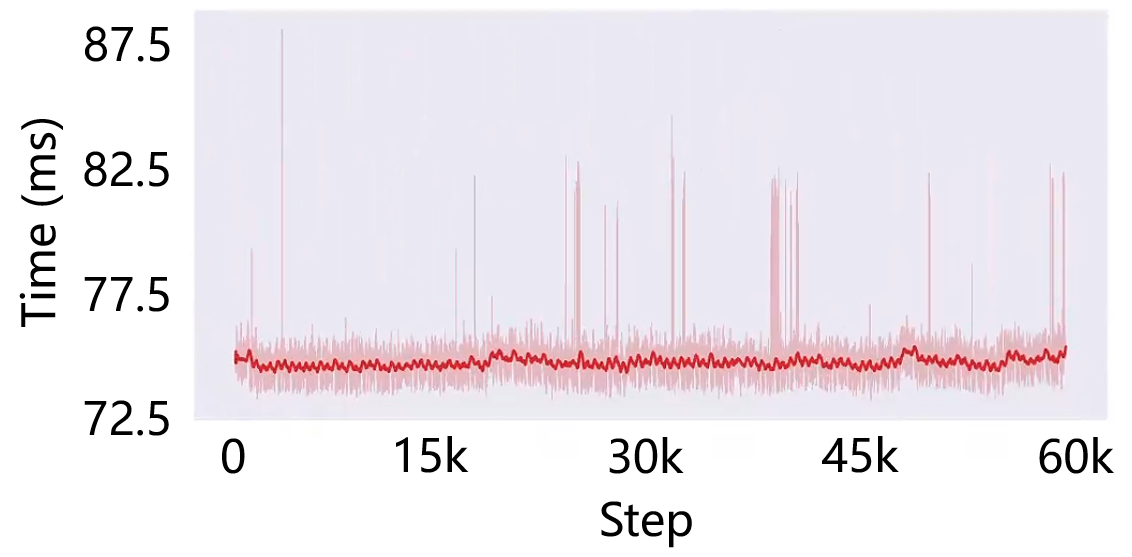}
        \caption{State perception time}
        \label{fig:state_perception_time}
    \end{subfigure}
    \hfill
    \begin{subfigure}{0.49\columnwidth}
        \centering
        \includegraphics[width=\linewidth]{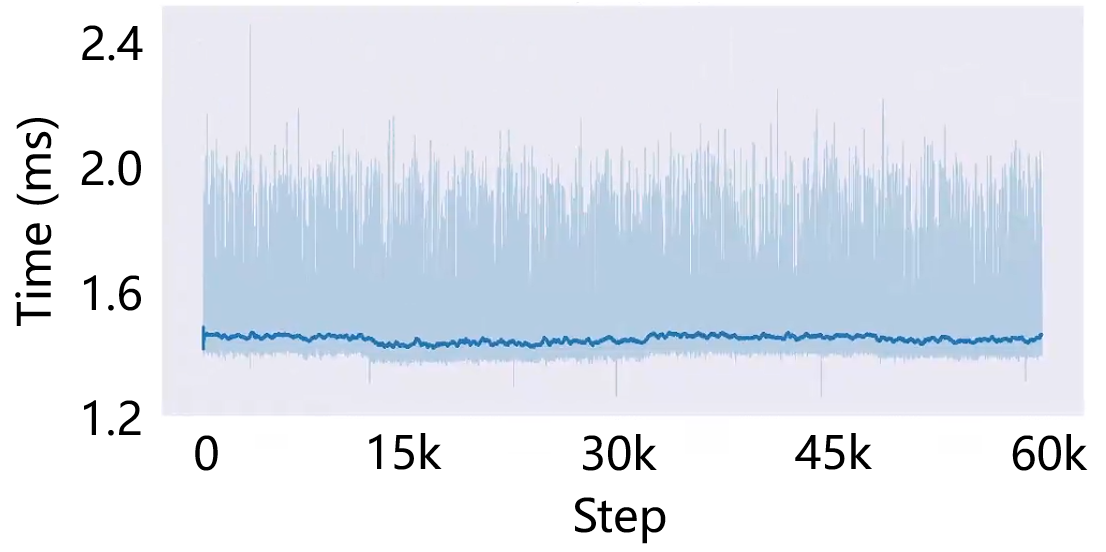}
        \caption{Policy inference time}
        \label{fig:policy_inference_time}
    \end{subfigure}

    \caption{Computational time analysis of the proposed framework during testing: (a) state perception time; (b) policy inference time. The light-colored curves represent the raw runtime, while the solid curves denote the smoothed runtime.}
    \label{fig:computational_time}
\end{figure}

Fig.~\ref{fig:computational_time} reports the single-step runtime of the state perception module and the policy inference module during testing. The runtime of the state perception module remains stable at around \(75\) ms, with an average time of \(74.96\) ms, a maximum time of \(89.96\) ms, and a minimum time of \(73.38\) ms. The policy inference module requires much less computation time, with an average runtime of \(1.44\) ms, a maximum runtime of \(2.44\) ms, and a minimum runtime of \(1.25\) ms. These results indicate that both the state perception and policy inference modules maintain low computational latency, suggesting the potential of the proposed framework for real-world deployment in autonomous racing.

\subsection{Discussion}

It is worth noting that the yaw-rate constraint in Eq. \eqref{equ:yaw rate constraint} explicitly depends on the longitudinal speed. 
As a result, the constraint becomes more restrictive during high-speed driving. If the constraint is satisfied mainly by reducing speed, dynamic-state violations can be reduced, but racing performance is also significantly compromised, as reflected by the conservative behavior of SAC-Lag. In contrast, the proposed method maintains a high average lap speed while keeping the number of constraint violations relatively low. 
This indicates that the introduced explicit dynamics constraints do not simply trade speed for safety. Instead, they guide the vehicle to operate efficiently near the boundary of the safe envelope, achieving a better balance between racing performance and dynamic stability.

As shown in Tables \ref{tab:results_testing} and \ref{tab:ablation_results}, the $\omega$-TaUT and $\beta$-TaUT are not reduced to zero. This is mainly because the proposed method adopts a Lagrangian-relaxation-based soft-constrained reinforcement learning mechanism, rather than hard constraint projection or an explicit safety filter. The stability costs can penalize constraint violations in the optimization objective, but they cannot guarantee strict constraint satisfaction at every time step. Meanwhile, although the sliding-window smoothing mechanism helps improve training stability, it may also weaken the direct suppression of instantaneous single-step violations. Therefore, TraD-RL should be more accurately interpreted as reducing and mitigating unsafe dynamic states through stability costs, rather than providing strict step-wise safety guarantees.

Regarding real-world deployment, although deep reinforcement learning methods generally suffer from the sim-to-real gap, the low-latency policy inference, structured low-dimensional observations, and stable performance under observation noise indicate that TraD-RL has certain potential for real-world deployment. Nevertheless, realistic physical factors such as tire--road friction variations, actuator dynamics, sensor delay, and model mismatch still require further validation in future work.

\section{Conclusion} \label{sec:conclusion}

This paper proposes TraD-RL, a reinforcement learning framework for autonomous racing that incorporates MCRL trajectory guidance, vehicle dynamics constraints, and two-stage curriculum learning into the training process. The MCRL provides global path and velocity references to improve exploration efficiency and racing performance. The yaw-rate and sideslip-angle constraints define the dynamic safe operating envelope and guide policy optimization through stability costs. The curriculum learning strategy enables a progressive transition from stable trajectory following to high-speed performance exploration. Experimental results on the Berlin and Modena racetracks show that TraD-RL achieves better lap performance than baseline methods including DDPG, PPO, TAL, and SAC-Lag, while maintaining a more reasonable balance between racing speed and dynamic stability. Ablation studies, parameter sensitivity analysis, and noise robustness tests further validate the effectiveness, generalization capability, and stability of the proposed method.

The current validation in this study is based on a simplified dynamic simulation environment and a linear tire model, which cannot fully capture nonlinear tire saturation characteristics under limit-handling conditions. Moreover, the present work focuses on single-agent autonomous racing with a fixed vehicle model and a fixed road-adhesion condition, without considering variations in vehicle parameters, road friction, adversarial behaviors, or multi-vehicle interactions. In future work, nonlinear tire models such as the Pacejka tire model and higher-fidelity vehicle dynamics simulation environments will be introduced. The effects of different vehicle parameter settings and varying road-adhesion conditions will be further investigated. In addition, action oscillations caused by direct neural-network policy outputs will be further addressed through action smoothing, control regularization, or safety-filter-based post-processing mechanisms. Adversarial and multi-vehicle racing scenarios will also be considered to more comprehensively evaluate the deployment potential of the proposed method under realistic competitive racing conditions.



\printcredits

\section*{Data availability}
The data and materials used to support the findings of this study are available from the corresponding author upon reasonable request.

\section*{Declaration of competing interest}
The authors declare that they have no known competing financial interests or personal relationships that could have appeared to influence the work reported in this paper.

\section*{Acknowledgments}
This work is supported by the National Natural Science Foundation of China under Grant No.52522219, and No.52325212.

\bibliographystyle{elsarticle-num}

\bibliography{cas-refs}

\bio{figs/authors/Lengbo}
\textbf{Bo Leng} received the Ph.D. degree in vehicle engineering from Tongji University, Shanghai, China. He is currently an Associate Professor with the School of Automotive Studies, Tongji University. His current research interests include the dynamic control of distributed drive electric vehicles and motion planning and control of intelligent vehicles. He has won the First Prize in the China Automobile Industry Technology Invention Award and the First Prize in the Shanghai Science and Technology Progress Awards in 2020 and 2022. He has been selected into the Young Elite Scientists Sponsorship Program of the China Association for Science and Technology in 2022.
\endbio

\bio{figs/authors/Zhangweiqi}
\textbf{Weiqi Zhang} received his B.E. degree in vehicle engineering from Tongji University, Shanghai, China, in 2025. Currently, he is working toward the Ph.D. degree at the Institute of Intelligent Vehicles, Tongji University, Shanghai, China. His research interests include autonomous racing, safe reinforcement learning, and end-to-end autonomous driving.
\endbio
\vspace{10pt}

\bio{figs/authors/Lizhuoren}
\textbf{Zhuoren Li} received his Ph.D. degree in vehicle engineering in 2025 and his B.E. degree in engineering mechanics in 2019, Tongji University, Shanghai, China. His current research interests include safe reinforcement learning, large language model enhanced end-to-end autonomous driving, interaction decision-making, and motion planning of autonomous vehicles.
\endbio
\vspace{10pt}

\bio{figs/authors/Xionglu}
\textbf{Lu Xiong} received the Ph.D. degree in vehicle engineering from Tongji University, Shanghai, China, in 2005. He is currently the Vice President and a Professor with the School of Automotive Studies, Tongji University. His current research interests include the dynamic control of distributed drive electric vehicles, motion planning and control of intelligent vehicles, and all-terrain vehicles. He won the First Prize in the Shanghai Science and Technology Progress Awards in 2013, 2020, and 2022. He was a recipient of the National Science Fund for Distinguished Young Scholars.
\endbio

\bio{figs/authors/Jinguizhe}
\textbf{Guizhe Jin} received the bachelor's degree in mechanical engineering from Harbin Institute of Technology, Weihai, China, in 2019. He is currently pursuing the master's degree with the College of Automotive and Energy Engineering, Tongji University, Shanghai, China. His research interests mainly focus on motion planning for autonomous vehicles via deep reinforcement learning.
\endbio

\bio{figs/authors/Yuran}
\textbf{Ran Yu} received his B.E. degree in vehicle engineering from the School of Automotive Engineering, Harbin Institute of Technology, Weihai, China, in 2024. He is currently pursuing his M.E. degree in vehicle engineering from the College of Automotive and Energy Engineering, Tongji University, Shanghai, China. His research interests include safe reinforcement learning, and motion planning and control of intelligent vehicles.
\endbio

\bio{figs/authors/Lvchen}
\textbf{Chen Lv}  received the joint Ph.D. degree from the Department of Automotive Engineering, Tsinghua University, China,
and the EECS Department, University of California, Berkeley, USA, in January 2016. He is an Associate Professor with the School of Mechanical and Aerospace Engineering, Nanyang Technological
University (NTU), Singapore. He also holds joint appointments as the Research Director of NTU, Vice President Research Office, the Director of M.Sc. degree in robotics and intelligent systems, a Thrust
Lead in AI with the Continental-NTU Corporation Laboratory, and the Cluster Director in Future Mobility Solutions with ERI@N. He was as a
Research Fellow with the Advanced Vehicle Engineering Center,  ranfield University, U.K., from 2016 to 2018. He joined NTU, as a Nanyang Assistant Professor and has founded the Automated Driving and Human-Machine System (AutoMan) Research Laboratory in June 2018. His research focuses on advanced vehicle control and intelligence, where he has contributed four books, over 200 papers, and received 12 granted patents and five patent applications.
\endbio
\vspace{10pt}

\end{document}